\documentclass[sigconf]{acmart}

\usepackage{graphicx}
\usepackage{amsmath}
\usepackage{epstopdf}
\usepackage{array}
\usepackage{multirow}
\usepackage{amsfonts}
\usepackage{booktabs}
\usepackage{subfigure}
\usepackage{bm}
\usepackage{algorithm}
\usepackage{algpseudocode}
\usepackage{color}
\usepackage{multirow}

\AtBeginDocument{%
  \providecommand\BibTeX{{%
	\normalfont B\kern-0.5em{\scshape i\kern-0.25em b}\kern-0.8em\TeX}}}

\copyrightyear{2020}
\acmYear{2020}
\setcopyright{acmcopyright}\acmConference[MM '20]{Proceedings of the 28th ACM International Conference on Multimedia}{October 12--16, 2020}{Seattle, WA, USA}
\acmBooktitle{Proceedings of the 28th ACM International Conference on Multimedia (MM '20), October 12--16, 2020, Seattle, WA, USA}
\acmPrice{15.00}
\acmDOI{10.1145/3394171.3413938}
\acmISBN{978-1-4503-7988-5/20/10}

\begin{document}
\fancyhead{}

\title{Efficient Crowd Counting via Structured Knowledge Transfer}

\author{Lingbo Liu}
\affiliation{%
  \institution{Sun Yat-Sen University}
}
\email{liulingb@mail2.sysu.edu.cn}

\author{Jiaqi Chen}
\affiliation{%
  \institution{Sun Yat-Sen University}
}
\email{jadgechen@gmail.com}

\author{Hefeng Wu}
\authornote{Hefeng Wu is the corresponding author. Lingbo Liu and Jiaqi Chen are co-first authors.}
\affiliation{%
  \institution{Sun Yat-Sen University}
}
\email{wuhefeng@gmail.com}

\author{Tianshui Chen}
\affiliation{%
  \institution{DarkMatter AI Research}
}
\email{tianshuichen@gmail.com}

\author{Guanbin Li}
\affiliation{%
  \institution{Sun Yat-Sen University}
}
\email{liguanbin@mail.sysu.edu.cn}

\author{Liang Lin}
\affiliation{%
  \institution{Sun Yat-Sen University}
  \institution{DarkMatter AI Research}
}
\email{linliang@ieee.org}

\renewcommand{\shortauthors}{Lingbo Liu, et al.}

\begin{abstract}
  Crowd counting is an application-oriented task and its inference efficiency is crucial for real-world applications. However, most previous works relied on heavy backbone networks and required prohibitive run-time consumption, which would seriously restrict their deployment scopes and cause poor scalability.
  To liberate these crowd counting models, we propose a novel Structured Knowledge Transfer (SKT) framework, which fully exploits the structured knowledge of a well-trained teacher network to generate a lightweight but still highly effective student network.
  Specifically, it is integrated with two complementary transfer modules, including an Intra-Layer Pattern Transfer which sequentially distills the knowledge embedded in layer-wise features of the teacher network to guide feature learning of the student network and an Inter-Layer Relation Transfer which densely distills the cross-layer correlation knowledge of the teacher to regularize the student's feature evolution.
  Consequently, our student network can derive the layer-wise and cross-layer knowledge from the teacher network to learn compact yet effective features.
  Extensive evaluations on three benchmarks well demonstrate the effectiveness of our SKT for extensive crowd counting models. In particular, only using around $6\%$ of the parameters and computation cost of original models, our distilled VGG-based models obtain at least 6.5$\times$ speed-up on an Nvidia 1080 GPU and even achieve state-of-the-art performance. Our code and models are available at {\color{blue}\url{https://github.com/HCPLab-SYSU/SKT}}.
\end{abstract}




\maketitle

\section{Introduction}

\begin{table}
\newcommand{\tabincell}[2]{\begin{tabular}{@{}#1@{}}#2\end{tabular}}
  \centering
  \resizebox{8.7cm}{!} {
    \begin{tabular}{c|c|c|c|c|c}
    \hline
    Method & RMSE &\#Param & FLOPs & GPU & CPU \\
    \hline\hline
    DISSNet~\cite{liu2019crowd}  & 159.20 & 8.86  & 8670.09 & 3677.98 & 378.80  \\
    CAN~\cite{liu2019context}    & 183.00 & 18.10 & 2594.18 & 972.16  & 149.56 \\
    CSRNet*~\cite{li2018csrnet}  & 233.32 & 16.26 & 2447.91 & 823.84  & 119.67 \\
    BL*~\cite{ma2019bayesian}    & 158.09 & 21.50 & 2441.23 & 595.72  & 130.76 \\
    \hline
    {\bf{Ours}}                        & {\color{red}\bf{156.82}} & {\color{red}\bf{1.35}} & {\color{red}\bf{155.30}}  & {\color{red}\bf{90.96}}  & {\color{red}\bf{9.78}} \\
    \hline
    \end{tabular}
  }
  \vspace{0mm}
   \caption{
   The Root Mean Squared Error (RMSE), Parameters, FLOPs, and inference time of our SKT network and four state-of-the-art models on the UCF-QNRF~\cite{idrees2018composition} dataset. The FLOPs and parameters are computed with the input size of 2032$\times$2912, and the inference times are measured on a Intel Xeon E5 CPU (2.4G) and a single Nvidia GTX 1080 GPU. The RMSE of the models with * are reimplemented by us. The units are million (M) for \#Param, giga (G) for FLOPs, millisecond (ms) for GPU time, and second (s) for CPU time, respectively. More efficiency analysis can be found in Table~\ref{tab:efficiency-all}.
   }
  \vspace{-5mm}
  \label{tab:efficiency-QNRF}
\end{table}

Crowd counting, whose objective is to automatically estimate the total number of people in surveillance scenes, is an important issue of crowd analysis~\cite{li2015crowded,zhan2008crowd}. With the rapid growth of urban population and the increasing demand for security analysis and early warning in large-scale crowded scenarios, this task has attracted extensive interest in academic and industrial fields, due to its wide-ranging applications in video surveillance~\cite{zhang2017fcn}, congestion alerting~\cite{semertzidis2010video} and traffic prediction~\cite{liu2018attentive,liu2020dynamic}.

Recently, deep neural networks \cite{zhang2015cross,boominathan2016crowdnet,zhang2016single,sindagi2017generating,liu2018crowd,cao2018scale,cheng2019improving,qiu2019crowd,liu2019adcrowdnet,zhang2019attentional,yuan2020crowd,ma2020learning} have become mainstream in the task of crowd counting and have made remarkable progress. 
To acquire better performance, most of the state-of-the-art methods \cite{li2018csrnet,liu2019context,ma2019bayesian,tan2019crowd,guo2019dadnet,liu2019crowd,yan2019perspective} utilized heavy backbone networks (such as the VGG model~\cite{simonyan2014very}) to extract features.
Nevertheless, requiring large computation cost and running at low speeds, these models are exceedingly inefficient. As shown in Table~\ref{tab:efficiency-QNRF}, the latest DISSNet~\cite{liu2019crowd} costs 3.7 seconds on an Nvidia 1080 GPU and 379 seconds on an Intel Xeon CPU to process an input image of size 2032$\times$2912.
This would seriously restrict their deployment scopes and cause poor scalability, particularly on edge computing devices~\cite{IOT} with limited computing resources. Moreover, to handle citywide surveillance videos in real-time, we may need thousands of high-performance GPUs, which are expensive and energy-consuming. Under these circumstances, a cost-effective model is extremely desired for crowd counting.

Thus, one fundamental question is that {\textit{how we can acquire an efficient crowd counting model from existing well-trained but heavy networks}.
A series of efforts~\cite{han2015deep,tai2015convolutional,cai2017deep,liu2018efficient,chen2018learning} have been made to compress and speed-up deep neural networks. However, most of them either require cumbersome hyper-parameters search (e.g., the sensitivity in per layer for parameters pruning) or rely on specific hardware platforms (e.g., weight quantization and half-precision floating-point computing).
Recently, knowledge distillation~\cite{hinton2015distilling}, which trains a small student network to acquire the knowledge of a complex teacher network, has become a desirable alternative for model compression due to its broad applicability areas. Numerous works~\cite{romero2014fitnets,zagoruyko2016paying,sau2016deep,chen2018fine,zhang2018deep,mirzadeh2019improved} have verified its effectiveness for image classification.
However, it is difficult to apply knowledge distillation to the more challenging dense-labeling crowd counting, since this task requires the distilled models to maintain the discriminative abilities at every location. In this case, extensive knowledge is desired to be transferred to well equip the abilities of student networks for generating accurate crowd density maps.

To fully distill the knowledge of a pre-trained teacher network, we conduct a more in-depth analysis of the crowd counting models and observe that the structured knowledge of deep networks is implicitly embedded in {\color{red}\textbf{i)}} {\textit{layer-wise features}} that carry image content, and {\color{red}\textbf{ii)}} {\textit{cross-layer correlations}} that encode feature updating schemas.
In this work, we develop a novel framework termed Structured Knowledge Transfer (SKT), which simultaneously distills the knowledge of layer-wise features and cross-layer correlation via two complementary transfer modules, i.e., an Intra-Layer Pattern Transfer (Intra-PT) and an Inter-Layer Relation Transfer (Inter-RT).
{\bf{First}}, our Intra-PT takes a set of representative features extracted from a well-trained teacher network to sequentially supervise the corresponding features of a student network, analogous to using the teacher's knowledge to progressively correct the student's learning deviation. As a result, the features of the student network exhibit similar visual or semantic patterns of its supervisor.
{\bf{Second}}, our Intra-PT densely computes the relationships between pairwise features of the teacher network and then utilizes such knowledge to help the student network regularize the long short-term evolution of its hierarchical features. Thereby, the student network can learn the solution procedure flow of its teacher.
Thanks to the tailor-designed SKT framework, our lightweight student network can effectively learn compact and knowledgeable features, yielding high-quality crowd density maps.

In experiments, we apply the proposed SKT framework to compress and accelerate a series of existing crowd counting models (e.g, CSRNet~\cite{li2018csrnet}, BL~\cite{ma2019bayesian} and SANet~\cite{cao2018scale}). Extensive evaluations on three representative benchmarks greatly demonstrate the effectiveness of our method. Under the condition of occupying only nearly 6\% of the original model parameters and computational cost, our distilled VGG-based models obtain at least 6.5$\times$ speed-up on GPU and 9$\times$ speed-up on CPU. Moreover, these lightweight models can preserve competitive performance, and even achieve state-of-the-art results on Shanghaitech\cite{zhang2016single} Part-A and UCF-QNRF \cite{idrees2018composition} datasets. 
In summary, the major contributions of this work are three-fold:
\begin{itemize}
\vspace{-1.0mm}
  \item We propose a general and comprehensive Structured Knowledge Transfer framework, which can generate lightweight but effective crowd counting models. To the best of our knowledge, we are the first to focus on improving the efficiency of crowd counting models.
  \item Two cooperative knowledge transfers (i.e., an Intra-Layer Pattern one and an Inter-Layer Relation one) are incorporated to fully distill the structured knowledge of well-trained models to lightweight models for crowd counting.
  \item Extensive experiments on three benchmarks show the effectiveness of our method. In particular, our distilled VGG-based models have an order of magnitude speed-up while achieving state-of-the-art performance.
\vspace{-1.75mm}
\end{itemize}

\section{Related Works}
{\bf{\large{Crowd Counting: }}}
Crowd counting has been extensively studied for decades. Early works~\cite{ge2009marked,li2008estimating} estimated the crowd count by directly locating the people with pedestrian detectors. Subsequently, some methods~\cite{Ryan2009multiple,chen2012feature} learned a mapping between handcrafted features and crowd count with regressors. Only using image low-level information, these methods had high efficiencies, but their performance was far from satisfactory for real-world applications.

\begin{figure*}[t]
\begin{center}
 \includegraphics[width=1.95\columnwidth]{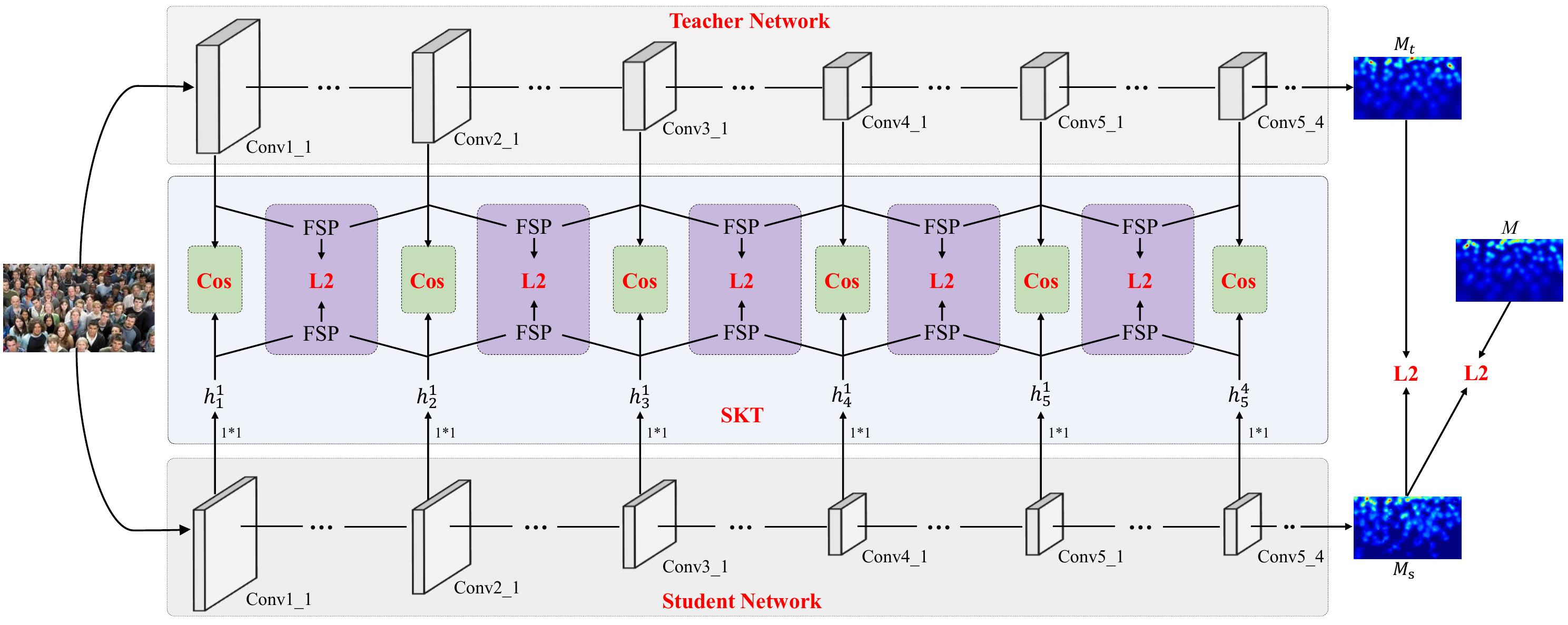}
 \vspace{-4mm}
\end{center}
   \caption{The proposed Structured Knowledge Transfer (SKT) framework for crowd counting.
   With two complementary distillation modules, our SKT can effectively distill the structured knowledge of a pre-trained teacher network to a small student network.
   First, an Intra-Layer Pattern Transfer sequentially distill the inherent knowledge in a teacher's feature to enhance a student's feature with a \textit{cosine} metric.
   Second, an Inter-Layer Relation Transfer enforces the student network to learn the long short term feature relationships of the teacher network, thereby fully mimicking the flow of solution procedure (FSP) of teacher.
   Notice that FSP matrices are densely computed between some representative features in our framework. For the conciseness and beautification of this figure, we only show some FSP matrices of adjacent features.
   }
\vspace{0mm}
\label{fig:model}
\end{figure*}

Recently, we have witnessed the great success of convolutional neural networks~\cite{zhang2015cross,onoro2016towards,walach2016learning,sam2017switching,liu2018crowd,shi2018crowd,shi2019revisiting,lian2019density,zhang2019attentional,sindagi2019multi,yan2019perspective} in crowd counting. Most of these previous approaches focused on how to improve the performance of deep models. To this end, they tended to use heavy backbone networks (e.g., the VGG model~\cite{simonyan2014very}) to extract representative features.
For instance, Li et al.~\cite{li2018csrnet} combined a VGG-16 based front-end network and dilated convolutional back-end network to learn hierarchical features for crowd counting.
Liu et al.~\cite{liu2019context} introduced an expanded context-aware network that learned both image features and geometry features with two truncated VGG16 models. Liu et al.~\cite{liu2019crowd} utilized three paralleled VGG16 networks to extract multiscale features and then conducted structured refinements. Recently, Ma et al.~\cite{ma2019bayesian} proposed a new Bayesian loss for crowd counting and verified its effectiveness on VGG19.
Although the aforementioned methods can make impressive progress, their performance advantages come with the cost of burdensome computation. Thus it is hard to directly apply these methods to practical applications.
In contrast, we take into consideration both the performance and computation cost. In this work, we aim to improve the efficiency of existing crowd counting models under the condition of preserving the performance.

{\bf{\large{Model Compression: }}}
Parameters quantization~\cite{gong2014compressing}, parameters pruning~\cite{han2015deep} and knowledge distillation~\cite{hinton2015distilling} are three types of commonly-used algorithms for model compression. Specifically, quantization methods~\cite{zhou2016dorefa,jacob2018quantization} compress networks by reducing the number of bits required to represent weights, but they usually rely on specific hardware platforms. Pruning methods~\cite{li2016pruning,he2017channel,zhu2017prune} removed redundant weights or channels of layers. However, most of them used weight masks to simulate the pruning and mass post-processing are needed to achieve real speed-up.
By contrast, knowledge distillation~\cite{sau2016deep,zhang2018deep,mirzadeh2019improved} is more general and its objective is to transfer knowledge from a heavy network to a lightweight one. Recently, it has been widely studied.
For instance, Hinton et al.~\cite{hinton2015distilling} trained a distilled network with the soft output of a large highly regularized network. Romero et al.~\cite{romero2014fitnets} improved the performance of student networks with both the outputs and the intermediate features of teacher networks. Zagoruyko and Komodakis~\cite{zagoruyko2016paying} utilized activation-based and gradient-based spatial attention maps to transfer knowledge between two networks. Nevertheless, most of these previous methods were proposed for image classification.
Recently, for fast human pose estimation, Zhang et al.~\cite{zhang2019fast} directly used the final pose maps of teacher networks to supervise the interim and final results of student networks. \cite{he2019knowledge} and \cite{liu2019structured} distilled the knowledge embedded in layer-wise features for semantic segmentation, but both of them neglected the the cross-layer correlation knowledge.
In contrast, our method fully explores the structured knowledge (e.g., layer-wise one and and cross-layer one) of teacher networks to optimize student networks for compact and effective feature learning.

\section{Method}
In this work, a general Structured Knowledge Transfer (SKT) framework is proposed to address the efficiency problem of existing crowd counting models. Its architecture is shown in Fig.~\ref{fig:model}. Specifically, an Intra-Layer Pattern Transfer (Intra-PT) and an Inter-Layer Relation Transfer (Inter-RT) are incorporated into our framework to fully transfer the structured knowledge (e.g., layer-wise one and cross-layer one) from the teacher networks to the student networks.

In this section, we take the VGG16-based CSRNet~\cite{li2018csrnet} as an example to introduce the working modules of our SKT framework. The student network is a $1$/$n$-CSRNet, in which the channel number of each convolutional layer (but except the last layer) is $1$/$n$ of the original one in CSRNet. Compared with the heavy CSRNet, the lightweight $1$/$n$-CSRNet model only has $1$/$n^2$ parameters and computation cost, but it suffers serious performance degradation.
Thus, our objective is to improve the performance of $1$/$n$-CSRNet as far as possible by transferring the knowledge of CSRNet.
Notice that our SKT is general and it is also applicable to other crowd counting models (e.g., BL~\cite{ma2019bayesian} and SANet~\cite{cao2018scale}). Several distilled models are analyzed and compared in Section~\ref{sec:experiment}.

\subsection{Feature Extraction}
In general, teacher networks have been pre-trained on standard benchmarks. The learned knowledge can be explicitly represented as parameters or implicitly embedded into features.
Similar to the previous works~\cite{romero2014fitnets,zagoruyko2016paying}, we perform knowledge transfer on the feature level. The knowledge in the features of teacher networks is treated as supervisory information and can be utilized to guide the representation learning of student networks. Therefore, before conducting knowledge transfer, we need to extract the hierarchical features of teacher/student networks in advance.

As shown in Fig.~\ref{fig:model}, given an unconstrained image $I$, we simultaneously feed it into CSRNet and $1$/$n$-CSRNet for feature extraction. For convenience, the $j$-th dilated convolutional layer in the back-end network of CSRNet is renamed as ``Conv$5$\_$j$'' in our work.
Thus the feature at layer Conv$i$\_$j$ of CSRNet can be uniformly denoted as $t_i^j$. Similarly, we use $s_i^j$ to represent the Conv$i$\_$j$ feature of $1$/$n$-CSRNet.
Notice that $t_i^j$ and $s_i^j$ share the same resolution, but $s_i^j$ has only $1$/$n$ channels of $t_i^j$.
Since the Inter-RT in Section~\ref{sec:Inter-RT} computes feature relations densely, we only perform distillation on some representative features, in order to reduce the computational cost during the training phase.
Specifically, the selected features are shown as follows:
{
\begin{equation}
T= \{t_1^1,~~ t_2^1,~~ t_3^1,~~~ t_4^1,~~~ t_5^1,~~ t_5^4\}, \quad
S= \{s_1^1,~~ s_2^1,~~ s_3^1,~~ s_4^1,~~ s_5^1,~~ s_5^4\},
\label{ms-ssim_loss}
\end{equation}
}%
where $T$ and $S$ are the feature groups of CSRNet and $1$/$n$-CSRNet, respectively.

\subsection{Intra-Layer Pattern Transfer}
As described in the above subsection, the extracted feature $t_i^j$ implicitly contains the learned knowledge of CSRNet. To improve the performance of the lightweight $1$/$n$-CSRNet, we design a simple but effective Intra-Layer Pattern Transfer (Intra-PT) module, which sequentially transfers the knowledge of selected features of CSRNet to the corresponding features of $1$/$n$-CSRNet. Formally, we enforce $s_i^j$ to learn the visual/semantic patterns of $t_i^j$ and optimize the parameters of $1$/$n$-CSRNet by maximizing their distribution similarity.

Specifically, our Intra-PT is composed of two steps. The {\bf{first}} step is channel adjustment. As feature $s_i^j$ and $t_i^j$ have different channel numbers, it is unsuitable to directly compute their similarity. To eliminate this issue, we generate a group of interim features $H= \{h_1^1,~ h_2^1,~ h_3^1,~ h_4^1,~ h_5^1,~ h_5^4\}$ by feeding each feature $s_i^j$ in $S$ into a $1\times1$ convolutional layer, which is expressed as:
{
\begin{equation}
h_i^j = s_i^j * w_{1\times1},\\
\end{equation}
}%
where $w_{1\times1}$ denotes the parameters of the convolutional layer. The output $h_i^j$ is the embedding feature of $s_i^j$ and its channel number is the same as that of $t_i^j$.

The {\bf{second}} step is similarity computation and knowledge transfer. Since Euclidean distance is too restrictive and may cause the rote learning of student networks, we adopt a relatively liberal metric, \textit{cosine}, to measure the similarity of two features. Specifically, the similarity of $t_i^j$ and $h_i^j$ at location $(x, y)$ is calculated by:
{
\setlength\abovedisplayskip{3pt}
\setlength\belowdisplayskip{3pt}
\begin{equation}
\begin{split}
\mathcal{S}_i^j(x,y)&= \textit{Cos}\{t_i^j(x,y), h_i^j(x,y)\}, \\
&= \sum_{c=1}^{\mathcal{C}_i^j} \frac{t_i^j(x,y,c)\cdot h_i^j(x,y,c)}{|t_i^j(x,y)|\cdot |h_i^j(x,y)|} ,
\end{split}
\end{equation}
}%
where $\mathcal{C}_i^j$ denotes the channel number of feature ${t_i^j}$ and $t_i^j(x,y,c)$ is the response value of ${t_i^j}$ at location $(x, y)$ of the $c$-th channel. The symbol $|\text{*}|$ is the length of a vector. Thus, the loss function of our Intra-PTD is defined as follows:
{
\begin{equation}
\mathcal{L}_{Intra}= \sum_{t_i^j, h_i^j} \sum_{x=1}^{\mathcal{H}_i^j} \sum_{y=1}^{\mathcal{W}_i^j} {1-\mathcal{S}_i^j(x,y)},
\label{uksd_loss}
\end{equation}
}%
where $\mathcal{H}_i^j$ and $\mathcal{W}_i^j$ are the height and width of feature $t_i^j$.
By minimizing this simple loss and back-propagating the gradients, our method can effectively transfer knowledge and optimize the parameters of $1$/$n$-CSRNet.
Compared with the pair-wise distillation~\cite{liu2019structured} which is heavy computed and can only conducted on high-layer features with low resolutions, our Intra-PT is more efficient and can fully transfer the knowledge embedded at various layers.

\subsection{Inter-Layer Relation Transfer}\label{sec:Inter-RT}
``Teaching one to fish is better than giving him fish''. Thus, a student network should also be encouraged to learn how to solve a problem. Inspired by \cite{yim2017gift}, the flow of solution procedure (FSP) can be modeled with the relationship between features from two layers. Such a relationship is a kind of meaningful knowledge.
In this subsection, we develop an Inter-Layer Relation Transfer (Inter-RT) module, which  densely computes the pairwise feature relationships (FSP matrices) of the teacher network to regularize the long short-term feature evolution of the student network .

Let's introduce the detail of our Inter-RT. We first present the generation of FSP matrix. For two general feature $f_1\in R^{h\times{w}\times{m}}$ and $f_2\in R^{h\times{w}\times{n}}$, we compute their FSP matrix $\mathcal{F}(f_1, f_2) \in R^{m\times n}$ with channel-wise inner product. Specifically, its value at index $(c_1,c_2)$ is calculated by:
{
\begin{equation}
\mathcal{F}_{c_1,c_2}(f_1, f_2) = \sum_{x=1}^h \sum_{y=1}^w \frac{f_1(x,y,c_1)\cdot f_2(x,y,c_2)}{h \cdot w}.
\label{fsp}
\end{equation}
}%
Notice that FSP matrix computation is conducted on features with same resolution. However, the features in $T$ have various resolutions. To address this issue and simultaneously reduce the FSP computation cost, we consistently resize all features in group $T$ to the resolution of $t_5^4$ with max pooling. The resized feature of $t_i^j$ is denoted as $R(t_i^j)$. In the same way, all features in group $H$ are also resized to the resolution of $h_5^4$.

Rather than only compute FSP matrices for adjacent features, we design a Dense FSP strategy to better capture the long-short term evolution of features.
Specifically, we generate a FSP matrix $\mathcal{F}\{R(t_i^j), R(t_k^l)\}$ for every pair of features $(t_i^j, t_k^l)$ in $T$. Similarly, a matrix $\mathcal{F}\{R(h_i^j), R(h_k^l)\}$ is also computed for every pair of features $(h_i^j, h_k^l)$ in $H$. Finally, the loss function of our Inter-RT is calculated as follows:
{
\begin{equation}
\mathcal{L}_{{Inter}}=\sum_{t_i^j, h_i^j}\sum_{t_k^l, h_k^l} || \mathcal{F}\{R(t_i^j), R(t_k^l)\} - \mathcal{F}\{R(h_i^j), R(h_k^l)\} ||^2.
\end{equation}
}%
By minimizing the distances of these FSP matrices, the knowledge of CSRNet can be transferred to $1$/$n$-CSRNet.

\begin{figure}[t]
\centering
   \includegraphics[width=0.95\columnwidth]{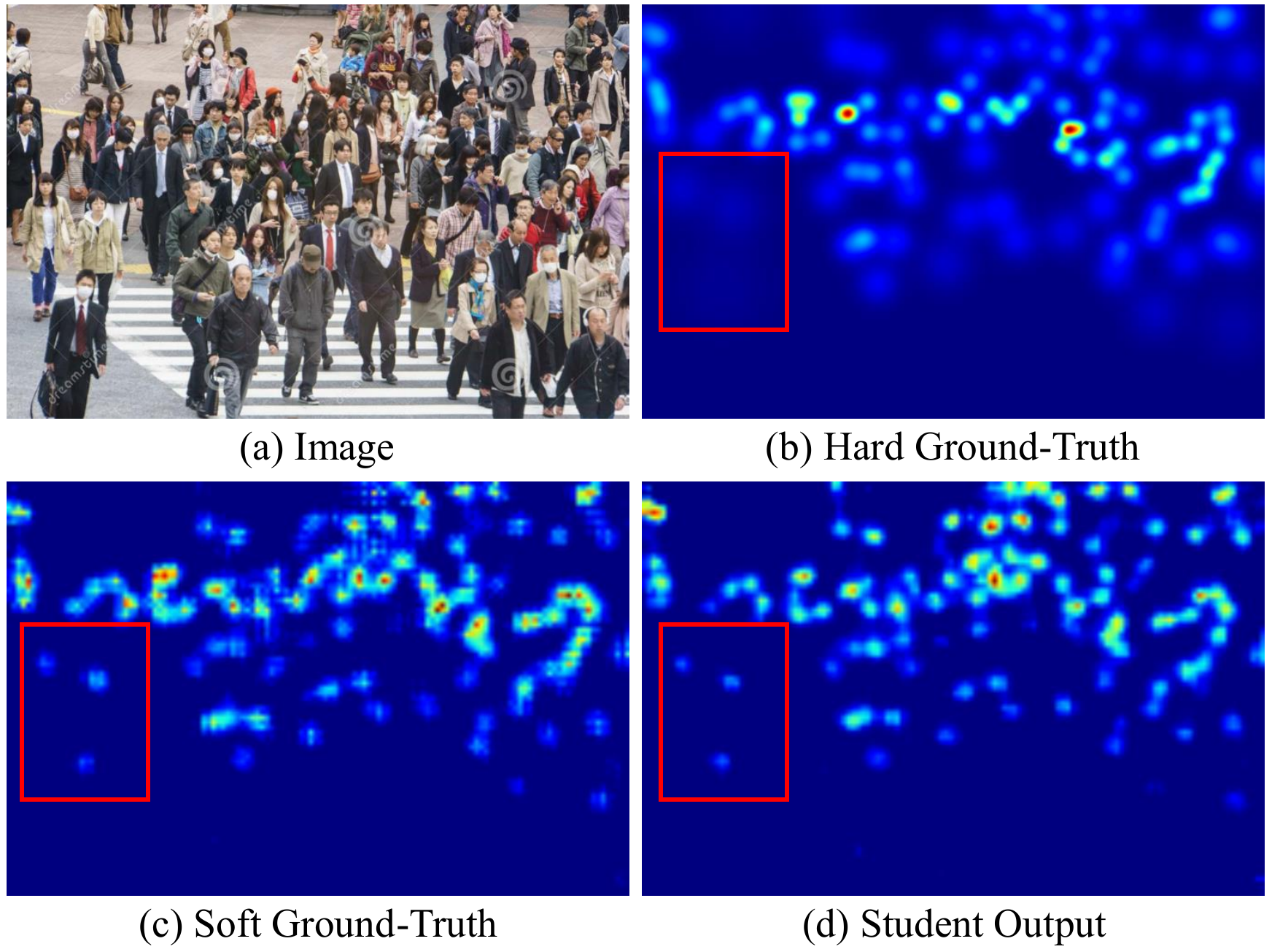}
\vspace{-2.5mm}
   \caption{Illustration the complementarity of hard and soft ground-truth (GT). (b) is a hard GT generated from point annotations with geometry-adaptive Gaussian kernels. (c) is a soft GT predicted by CSRNet, while (d) is the estimated map of $1$/$n$-CSRNet.
   The hard GT may exist some blemishes (e.g., the inaccurate scales and positions of human heads, the unmarked heads). For example, the red box in (b) shows the human heads with inaccurate scales.
   We find that the soft GT may be relatively reasonable in some regions and it is complementary to the hard GT.
   Thus, they can be incorporated to train the student network.}
\vspace{0mm}
\label{fig:soft-GT}
\end{figure}

\subsection{Learn from Soft Ground-Truth}
In our work, a density map generated from point annotations is termed as hard ground-truth. We find that the density maps predicted by the teacher network are complementary to hard ground-truths. As shown in Fig.~\ref{fig:soft-GT}, there may exist some blemishes (e.g., the inaccurate scales and positions of human heads, the unmarked heads) in some regions of hard ground-truths.
Fortunately, with powerful knowledge, a well-trained teacher network may predict some relatively reasonable maps. These predicted density maps can also be treated as knowledge and we call them soft ground-truths. In this work, we train our student network with both the hard and soft ground-truths.

As shown in Fig.~\ref{fig:model}, we use $M$ to represent the hard ground-truth of image $I$. The predicted map of CSRNet is denoted as $M_t$ and the output map of $1$/$n$-CSRNet is denoted as $M_s$. Since $1$/$n$-CSRNet is expected to simultaneously learn the knowledge of hard ground-truth and soft ground-truth, we defined the loss function on density maps as follows:
\begin{equation}
\mathcal{L}_m= || M_s - M||^2+|| M_s - M_t||^2.
\end{equation}
Finally, we optimize the parameters of 1/n-CSRNet by minimizing the total losses of all knowledge transfers:
\begin{equation}
\mathcal{L}= \alpha_{1}\mathcal\cdot{L}_{Intra} + \alpha_{2}\cdot\mathcal{L}_{Inter} + \alpha_3\cdot\mathcal{L}_m, 
\label{final_loss}
\end{equation}
where $\alpha_{1}$, $\alpha_{2}$ and $\alpha_3$ are weights of different losses.

\section{Experiments}\label{sec:experiment}


\subsection{Experiment Settings}

In this work, we conduct extensive experiments on the following three public benchmarks for crowd counting.

{\bf{Shanghaitech \cite{zhang2016single}:}} This dataset contains 1,198 images with 330,165 annotated individuals. It is composed of two parts: Part-A contains 482 images of congested crowd scenes, where 300 images are used for training and 182 for testing, and Part-B contains 716 images of sparse crowd scenes, with 400 images for training and the rest for testing.

{\bf{UCF-QNRF \cite{idrees2018composition}:}} As one of the most challenging datasets, UCF-QNRF contains 1,535 images captured from unconstrained crowd scenes with huge variations in scale, density and viewpoint. Specifically, 1,201 images are used for training and 334 for testing. There are about 1.25 million annotated people in this dataset and the number of persons per image varies from 49 to 12,865.

{\bf{WorldExpo'10 \cite{zhang2015cross}:}} It contains 1,132 surveillance videos captured by 108 cameras during the Shanghai WorldExpo 2010. Specifically, 3,380 images from 103 scenes are used as the training set and 600 images from other five scenes as the test set. Region-of-Interest (ROI) are provided to specify the counting regions for the test set.

Following \cite{cao2018scale,li2018csrnet}, we adopt Mean Absolute Error (MAE) and Root Mean Squared Error (RMSE) to quantitatively evaluate the performance of crowd counting. Specifically, they are defined as follows:
{
\small
\setlength\abovedisplayskip{3pt}
\setlength\belowdisplayskip{3pt}
\begin{equation}
  \text{MAE} = \frac{1}{N}\sum_{i=1}^N\|P_i-G_i\|,
  \text{RMSE} = \sqrt{\frac{1}{N}\sum_{i=1}^N\|P_i-G_i\|^2},
\end{equation}
}%
where $N$ is the number of test images, $P_i$ and $G_i$ are the predicted and ground-truth count of $i^{th}$ image, respectively.

\begin{table}
\newcommand{\tabincell}[2]{\begin{tabular}{@{}#1@{}}#2\end{tabular}}
  \centering

    \begin{tabular}{c|c|c|c|c}
    \hline
    CPR  & \#Param & FLOPs & RMSE  & Transfer  \\
    \hline
    \hline
    1 & 16.26 & 205.88 & 105.99 &  \\
    \hline
    \multirow{2}{*}{$1$/$2$}& \multirow{2}{*}{4.07} & \multirow{2}{*}{51.77} &	137.32 &  \\
    \cline{4-4}
     & & & {\bf\color{red}113.61} & \checkmark \\
    \hline
    \multirow{2}{*}{$1$/$3$} & \multirow{2}{*}{1.81} & \multirow{2}{*}{23.11} & 140.29 &  \\
    \cline{4-4}
    & &  &  {\bf\color{red}114.68} & \checkmark \\
    \hline
    \multirow{2}{*}{$1$/$4$} & \multirow{2}{*}{1.02} & \multirow{2}{*}{13.09} & 146.40 & \\
    \cline{4-4}
    &  & & {\bf\color{red}114.40} & \checkmark \\
    \hline
    \multirow{2}{*}{$1$/$5$} & \multirow{2}{*}{0.64} & \multirow{2}{*}{8.45} & 149.40 & \\
    \cline{4-4}
    & & &  {\bf\color{red}118.78} & \checkmark \\
    \hline
    \end{tabular}

  \vspace{0mm}
   \caption{Performance under different channel preservation rates (CPR) on Shanghaitech Part-A. \#Param denotes the number of parameters (M). FLOPs is the number of FLoating point OPerations (G) and it is computed on a $576 \times 864$ image ($576 \times 864$ is the average resolution on Shanghaitech Part-A).
   }
  \vspace{-4mm}
  \label{tab:Distillation-Rate}
\end{table}

\begin{figure*}[t]
\begin{center}
 \includegraphics[width=1.9\columnwidth]{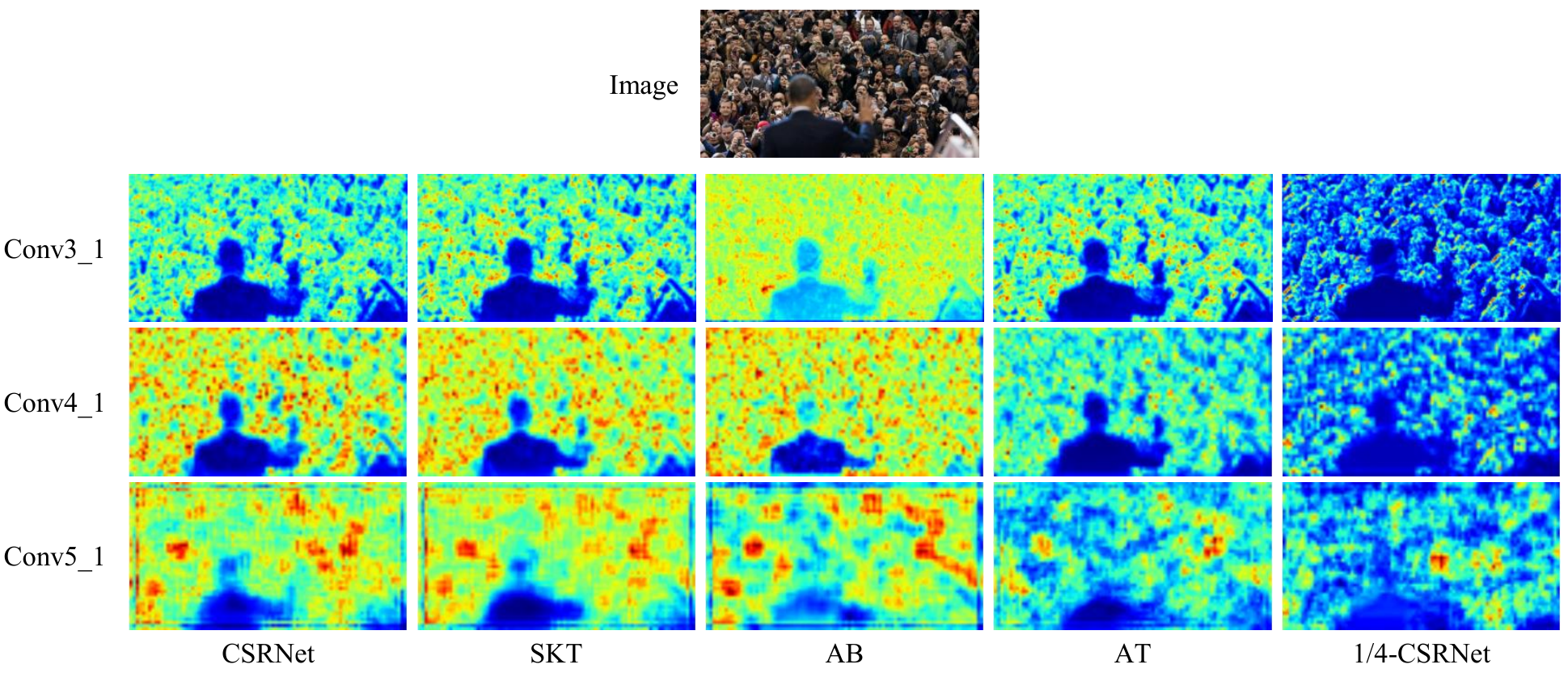}
 \vspace{-5mm}
\end{center}
   \caption{Visualization of the feature maps of different models on Shanghaitech Part-A.
   The first and fifth columns are the features of the complete CSRNet and the naive 1/4-CSRNet. The middle three columns show the student features of 1/4-CSRNet+SKT, 1/4-CSRNet+AB~\cite{heo2019knowledge} and 1/4-CSRNet+AT~\cite{zagoruyko2016paying}.
   The bottom three rows are the channel-wise average features at layers Conv3\_1, Conv4\_1 and Conv5\_1, respectively.
   Thanks to the tailor-designed Intra-PT and Inter-RT, our 1/4-CSRNet+SKT can fully absorb the structured knowledge of CSRNet, thus the generated features are very similar to those of the teacher network.}
\vspace{0mm}
\label{fig:feature_visual}
\end{figure*}

\subsection{Ablation Study}
\subsubsection{\bf{Exploration on Channel Preservation Rate}}
In this work, we compress the existing crowd counting models by reducing their channel numbers. A model can run more efficiently if it has fewer parameters/channels, however, at the expense of performance. Thus, the balance between efficiency and accuracy should be investigated. In this section, we first conduct ablation study to evaluate the influence of Channel Preservation Rate (CPR) on the performance of models.

In Table \ref{tab:Distillation-Rate}, we summarize the performance of various CSRNet trained with different CPRs. As can be observed, the original CSRNet has 16.26M parameters and achieves an RMSE of 105.99 using 205.88G FLOPs. When preserving half the number of channels, 1/2-CSRNet has a 4$\times$ reduction in parameters and FLOPs. However, without applying knowledge transfer, its performance seriously degrades, with the RMSE increasing to 137.32. By applying our SKT, 1/2-CSRNet exhibits an obvious performance gain. When CPR decreases to 1/4, we can observe the model is further reduced in model size and FLOP consumption with only 6.25\% of the original one. What's more, the 1/4-CSRNet with knowledge transfer just has a negligible performance drop. When CPR further decreases, 1/5-CSRNet meets a relatively large performance drop with a smaller gain in parameters and FLOPs reduction. Therefore, we consider it roughly reaches a balance when CPR is 1/4 and this setting is widely adopted in the following experiments.

\begin{table}
\newcommand{\tabincell}[2]{\begin{tabular}{@{}#1@{}}#2\end{tabular}}
  \centering
    \begin{tabular}{c|c|c|c}
    \hline
    \multicolumn{2}{c|}{Transfer Configuration}  & MAE  & RMSE \\
    \hline
    \hline
    \multicolumn{2}{c|}{W/O Transfer}   & 89.65 & 146.40 \\
    \hline
    \multirow{2}{*}{Intra-PT} & L2 & 76.61 & 120.56 \\
    \cline{2-4}
    & Cos  & 74.99 & 117.58 \\
    \hline
    \multirow{2}{*}{Inter-RT} & S-FSP & 79.22 & 133.21 \\
    \cline{2-4}
    & D-FSP  & 73.25 & 120.77 \\
    \hline
    \multirow{2}{*}{Intra-PT \& Inter-RT} & L2 + D-FSP & 72.89 & 117.92 \\
    \cline{2-4}
    & Cos + D-FSP   & 71.55 & 114.40 \\
    \hline
    \end{tabular}
  \vspace{0mm}
   \caption{Performance of $1$/$4$-CSRNet distilled with different transfer configurations on Shanghaitech Part-A. D-FSP and S-FSP refer to Dense FSP and Sparse FSP, respectively.
   }
  \vspace{-4mm}
  \label{tab:KD-Term}
\end{table}

\begin{table}
\newcommand{\tabincell}[2]{\begin{tabular}{@{}#1@{}}#2\end{tabular}}
  \centering
    \begin{tabular}{c|c|c}
    \hline
   Ground-Truth Type  &  MAE & RMSE \\
    \hline
    \hline
    Hard &  72.94  & 116.68 \\
    \hline
    Soft &  74.89 & 118.33 \\
    \hline
    Hard + Soft &  71.55 & 114.40   \\
    \hline
    \end{tabular}
  \vspace{0mm}
   \caption{Performance of $1$/$4$-CSRNet trained with different ground-truths on Shanghaitech Part-A.
   }
  \vspace{-2mm}
  \label{tab:soft-GT}
\end{table}

\begin{table}
  \centering
    \begin{tabular}{c|c|c|c}
    \hline
    \multicolumn{2}{c|}{Method} & MAE & RMSE \\
    \hline\hline
    \multirow{2}{*}{Baseline} & CSRNet & 68.43 & 105.99 \\
                       & $1$/$4$-CSRNet & 89.65 & 146.40\\
    \hline
    \multirow{2}{*}{Quantization} & DoReFa~\cite{zhou2016dorefa} & 80.02 & 124.1 \\
                                  & QAT~\cite{jacob2018quantization} & 75.50 & 128.09 \\
    \hline
    \multirow{3}{*}{Pruning} & L1Filter~\cite{li2016pruning} & 85.18 & 135.82 \\
                             & CP~\cite{he2017channel} & 82.05 & 130.65 \\
                             & AGP~\cite{zhu2017prune} & 78.51 & 125.83 \\
    \hline
    \multirow{6}{*}{Distillation}
     & FitNets~\cite{romero2014fitnets} & 87.32 & 140.34 \\
     & DML~\cite{zhang2018deep} & 85.23 & 138.10 \\
     & NST~\cite{huang2017like} & 76.26 & 116.57\\
     & AT~\cite{zagoruyko2016paying} & 74.65 & 127.06\\
     & AB~\cite{heo2019knowledge} & 75.73 & 123.28\\
     & SKT (Ours) & {\bf\color{red}71.55} & {\bf\color{red}114.40}  \\
    \hline
    \end{tabular}
  \vspace{0mm}
   \caption{Performance of different compression algorithms on Shanghaitech Part-A.
   }
  \vspace{-5mm}
  \label{tab:compression_models}
\end{table}

\subsubsection{\bf{Effect of Different Transfer Configurations}}

We further perform experiments to evaluate the effect of different transfer configurations of our framework. This ablation study is conducted based on 1/4-CSRNet and the results are summarized in Table~\ref{tab:KD-Term}.
When just trained with our Intra-Layer Pattern Transfer module, the 1/4-CSRNet is able to obtain evident performance gain, decreasing MAE by at least 13 and RMSE by at least 25.8. We observe that using the Cosine (Cos) as the similarity metric performs better than using Euclidian (L2) distance. The possible reason is that the Cosine metric enforces the consistency of feature distribution between teacher and student networks, while the Euclidean distance further enforces location-wise similarity which is too restrictive for knowledge transfer.
On the other hand, only using our Inter-Layer Relation Transfer module can also boost the 1/4-CSRNet's performance by a large margin, with MAE decreased by at least 10 and RMSE decreased by at least 13. It is worth noting that the Dense FSP strategy achieves quite impressive performance gain, decreasing MAE of the 1/4-CSRNet without transfer by 16.40 (relatively 18.3\%).
When combining both the proposed Intra-Layer and Intra-Layer transfers to form our overall framework, the 1/4-CSRNet's performance is further boosted. Specifically, by using Cosine metric and Dense FSP, the 1/4-CSRNet achieves the best performance (MAE 71.55, RMSE 114.40) within all transfer configurations of our framework.

\begin{table}
  \centering
    \begin{tabular}{c|c|c|c|c}
    \hline
    \multirow{2}{*}{Method} &
    \multicolumn{2}{c|}{Part-A} &
    \multicolumn{2}{c}{Part-B} \\
    \cline{2-5}
    & MAE  & RMSE & MAE & RMSE \\
    \hline\hline
    MCNN~\cite{zhang2016single} & 110.2 & 173.2 & 26.4 & 41.3\\
    SwitchCNN~\cite{sam2017switching} & 90.4 & 135 & 21.6 & 33.4\\
    DecideNet~\cite{liu2018decidenet} & - & - & 21.5 & 31.9 \\
    CP-CNN~\cite{sindagi2017generating} & 73.6 & 106.4 & 20.1 & 30.1\\
    DNCL~\cite{shi2018crowd} & 73.5 & 112.3 & 18.7 & 26.0\\
    ACSCP~\cite{shen2018crowd} & 75.7 & 102.7 & 17.2 & 27.4 \\
    L2R~\cite{liu2018leveraging} & 73.6 & 112.0 & 13.7 & 21.4 \\
    IG-CNN~\cite{babu2018divide} & 72.5 & 118.2 & 13.6 & 21.1 \\
    IC-CNN~\cite{ranjan2018iterative} & 68.5 & 116.2 & 10.7 & 16.0\\
    CFF~\cite{shi2019counting} & 65.2 & 109.4 & 7.2 & 12.2\\
    \hline
    SANet* & 75.33 & 122.2 & 10.45 & 17.92 \\
    $1$/$4$-SANet & 97.36 & 155.43 & 14.79 & 23.43 \\
    $1$/$4$-SANet+SKT & {\bf\color{red}78.02} & {\bf\color{red}126.58} & {\bf\color{red}11.86} & {\bf\color{red}19.83} \\
    \hline
    CSRNet* & 68.43 & 105.99 & 7.49 & 12.33 \\
    $1$/$4$-CSRNet & 89.65 &	146.40 & 10.82 & 16.21 \\
    $1$/$4$-CSRNet + SKT & {\bf\color{red}71.55} & {\bf\color{red}114.40} & {\bf\color{red}7.48} & {\bf\color{red}11.68} \\
    \hline
    BL* &  61.46 & 103.17 & 7.50 & 12.60 \\
    $1$/$4$-BL & 88.35 & 145.47 & 12.25 & 19.77 \\
    $1$/$4$-BL + SKT & {\bf\color{red}62.73} & {\bf\color{red}102.33} & {\bf\color{red}7.98} & {\bf\color{red}13.13}  \\
    \hline
    \end{tabular}
  \vspace{0mm}
   \caption{Performance comparison on Shanghaitech dataset. The models with symbol * are our reimplemented teacher networks.
   }
  \vspace{-6mm}
  \label{tab:ShanghaiTech}
\end{table}

\subsubsection{\bf{Effect of Soft Ground-Truth}}
In this section, we conduct experiments to evaluate the effect of soft ground-truth (GT) on the performance. As shown in Table~\ref{tab:soft-GT}, when only using the soft GT generated by the teacher network as supervision, the 1/4-CSRNet's performance is slightly worse than that of using hard GT. But it also indicates that the soft GT does provide useful information since it does not cause severe performance degradation. Furthermore, it is witnessed that the model's performance is promoted when we utilize both soft GT and hard GT to supervise model training. This further demonstrates that the soft GT is complementary to the hard GT and we can indeed transfer knowledge of the teacher network with soft GT.

\subsection{Comparison with Model Compression Algorithms}
Undoubtedly, some existing compression algorithms can also be applied to compress the crowd counting models. To verify the superiority of the proposed SKT, we compare our method with ten representative compression algorithms.

In Table \ref{tab:compression_models}, we summarize the performance of different compression algorithms on Shanghaitech Part-A.
Specifically, quantizing the parameters of CSRNet with 8 bits, DoReFa~\cite{zhou2016dorefa} and QAT~\cite{jacob2018quantization} obtain an MAE of 80.02/75.50 respectively. When we employ the official setting of CP~\cite{he2017channel} to prune CSRNet, the compressed model obtains an MAE of 82.05 with 6.89M parameters. To maintain the same number of parameters of 1/4-CSRNet, L1Filter~\cite{li2016pruning} and AGP~\cite{zhu2017prune} prunes 93.75\% parameters, and their MAE are above 78.
Furthermore, six distillation methods including our SKT are applied to distill CSRNet to 1/4-CSRNet. As can be observed, our method achieves the best performance w.r.t both MAE and RMSE. The feature visualization in Fig.~\ref{fig:feature_visual} also shows that our features are much better than those of other compression methods. These
quantitative and qualitative superiorities are attributed to that the tailor-designed Intra-PT and Inter-RT can fully distill the knowledge of the teacher networks.
What's more, the proposed SKT is easily implemented and the distilled crowd counting models can be directly deployed on various edge devices.
In summary, our SKT fits most with the crowd counting task, among the various existing compression algorithms.

\begin{table}
\newcommand{\tabincell}[2]{\begin{tabular}{@{}#1@{}}#2\end{tabular}}
  \centering
    \begin{tabular}{c|c|c}
    \hline
    Method & MAE & RMSE\\
    \hline
    \hline
    Idrees et al. ~\cite{idrees2013multi}           & 315 & 508 \\
    MCNN~\cite{zhang2016single}                     & 277 & 426 \\
    Encoder-Decoder~\cite{badrinarayanan2015segnet} & 270 & 478 \\
    CMTL~\cite{sindagi2017cnn}                      & 252 & 514 \\
    SwitchCNN~\cite{sam2017switching}               & 228 & 445 \\
    Resnet-101~\cite{he2016deep}                    & 190 & 277 \\
    CL~\cite{idrees2018composition}                 & 132 & 191 \\
    TEDnet~\cite{jiang2019crowd} & 113      & 188 \\
    CAN~\cite{liu2019context}    & 107	    & 183 \\
    S-DCNet~\cite{xiong2019open} & 104.40	& 176.10 \\
    DSSINet~\cite{liu2019crowd} & 99.10	& 159.20 \\
    \hline
    SANet* & 152.59 & 246.98\\
    $1$/$4$-SANet & 192.47 & 293.96\\
    $1$/$4$-SANet + SKT & {\bf\color{red}157.46} & {\bf\color{red}257.66}\\
    \hline
    CSRNet* & 145.54 & 233.32  \\
    $1$/$4$-CSRNet & 186.31 &	287.65  \\
    $1$/$4$-CSRNet + SKT & {\bf\color{red}144.36} & {\bf\color{red}234.64} \\
    \hline
    BL* & 87.70 & 158.09  \\
    $1$/$4$-BL{~~~~~~~~~~~~} & 135.64 &	224.72  \\
    $1$/$4$-BL + SKT & {\bf\color{red}96.24} & {\bf\color{red}156.82} \\
    \hline
    \end{tabular}
  \vspace{0mm}
    \caption{Performance of different methods on UCF-QNRF dataset. The models with * are our reimplemented teacher networks.
    }
  \vspace{-4mm}
  \label{tab:UCF-QNRF}
\end{table}
\subsection{Comparison with Crowd Counting Methods}

To demonstrate the effectiveness of the proposed SKT, we also conduct comparisons with state-of-the-art methods of crowd counting from both performance and efficiency perspectives. Besides CSRNet \cite{li2018csrnet}, we also apply our SKT framework to distill other two representative models BL \cite{ma2019bayesian} and SANet~\cite{cao2018scale}.
Specifically, the former is based on VGG19 and we obtain a lightweight 1/4-BL with the same transfer configuration of CSRNet. Similar to GoogLeNet \cite{szegedy2015going}, the latter SANet adopts multi-column blocks to extract features. For SANet, we transfer knowledge on the output features of each block, yielding a lightweight 1/4-SANet.

\begin{table*}[t]
\newcommand{\tabincell}[2]{\begin{tabular}{@{}#1@{}}#2\end{tabular}}
  \centering
  \resizebox{17.95cm}{!} {
    \begin{tabular}{c|c|c|c|c||c|c|c||c|c|c||c|c|c}
    \hline
    \multirow{2}{*}{Method} & \multirow{2}{*}{\#Param} &
    \multicolumn{3}{c||}{Shanghaitech A (576$\times$864)} & \multicolumn{3}{c||}{Shanghaitech B (768$\times$1024)} & \multicolumn{3}{c||}{WorldExpo'10 (576$\times$720)} & \multicolumn{3}{c}{UCF-QNRF (2032$\times$2912)} \\
    \cline{3-14}
    & & FLOPs & GPU  & CPU & FLOPs & GPU  & CPU & FLOPs & GPU  & CPU & FLOPs & GPU  & CPU \\
    \hline\hline
DSSINet~\cite{liu2019crowd}  &	8.86 &	729.20 &	296.32 &	32.39 &	1152.31 & 	471.83 &	49.49 &	607.66 &	250.69 & 26.13 &	8670.09 & 3677.98	&	378.80 \\
CAN~\cite{liu2019context} &	18.10&	218.20&	79.02&	7.99&	344.80&	117.12&	20.75&	181.83&	68.00&	6.84&	2594.18&	972.16&	149.56\\
CSRNet~\cite{li2018csrnet} &	16.26&	205.88&	66.58&	7.85&	325.34&	98.68&	19.17&	171.57&	57.57&	6.51&	2447.91&	823.84&	119.67\\
BL~\cite{ma2019bayesian} &	21.50&	205.32&	47.89&	8.84&	324.46&	70.18&	19.63&	171.10&	40.52&	6.69&	2441.23&	595.72&	130.76\\
SANet~\cite{cao2018scale} &	0.91&	33.55&	35.20&	3.90&	52.96&	52.85&	11.42&	27.97&	29.84&	3.13&	397.50&	636.48&	87.50\\
    \hline
    $1$/$4$-CSRNet + SKT & 1.02&	13.09&	8.88&	0.87&	20.69&	12.65&	1.84&	10.91&	7.71&	0.67&	155.69&	106.08&	9.71 \\
    $1$/$4$-BL + SKT & 1.35&	13.06&	7.40&	0.88&	20.64&	10.42&	1.89&	~~10.88&	6.25&	0.69&	155.30&	90.96&	9.78 \\
    $1$/$4$-SANet + SKT & 0.058 & 2.52 & 11.83 & 1.10 & 3.98 & 16.86 & 2.10 &  2.10 & 9.72 & 0.92 & 29.92 & 368.04 &	18.64\\
    \hline
    \end{tabular}
  }
  \vspace{0mm}
   \caption{The inference efficiency of state-of-the-art methods. \#Param denotes the number of parameters, while FLOPs is the number of FLoating point OPerations. The execution time is computed on an Nvidia GTX 1080 GPU and a 2.4 GHz Intel Xeon E5 CPU. The units are million (M) for \#Param, giga (G) for FLOPs, millisecond (ms) for GPU time, and second (s) for CPU time.
   }
  \vspace{-2mm}
  \label{tab:efficiency-all}
\end{table*}

\begin{table}
\centering
\newcommand{\tabincell}[2]{\begin{tabular}{@{}#1@{}}#2\end{tabular}}
  \centering
  \resizebox{8.0cm}{!} {
    \begin{tabular}{c|c|c|c|c|c|c}
    \hline
    Method & S1 & S2 & S3 & S4 & S5 & Avg\\
    \hline\hline
    Zhang et al~\cite{zhang2015cross}          & 9.8	&   14.1 &  14.3 &  22.2 & 3.7	& 12.9\\
    MCNN~\cite{zhang2016single}                & 3.4	&   20.6 &	12.9 &	13.0 & 8.1	& 11.6\\
    Shang et al.~\cite{shang2016end}           & 7.8	&   15.4 &  14.9 &	11.8 & 5.8	& 11.7\\
    IG-CNN~\cite{babu2018divide}               & 2.6    &   16.1 &  10.1 &  20.2 & 7.6  & 11.3 \\
    ConvLSTM~\cite{xiong2017spatiotemporal}    & 7.1	&   15.2 &	15.2 &	13.9 & 3.5	& 10.9\\
    IC-CNN~\cite{ranjan2018iterative}          & 17.0   &   12.3 &   9.2 &   8.1 & 4.7  & 10.3 \\
    SwitchCNN~\cite{sam2017switching}          & 4.4	&   15.7 &	10.0 &  11.0 & 5.9	& 9.4\\
    DecideNet~\cite{liu2018decidenet}          & 2.00   &  13.14 &  8.90 & 17.40 & 4.75 & 9.23 \\
    DNCL~\cite{shi2018crowd}                   & 1.9    &   12.1 &  20.7 &   8.3 & 2.6  & 9.1 \\
    CP-CNN~\cite{sindagi2017generating}        & 2.9	&   14.7 &	10.5 &	10.4 & 5.8	& 8.86\\
    PGCNet~\cite{yan2019perspective}        & 2.5   &   12.7   &   8.4   &   13.7   &   3.2   &   8.1 \\
    TEDnet~\cite{jiang2019crowd}        & 2.3   &   10.1   &   11.3  &   13.8   &   2.6   &   8.0 \\
    \hline
    SANet*         & 2.92 &	15.22 &	14.86 &	14.73 &	4.20 & 10.39\\
    $1$/$4$-SANet  & 3.77 &	19.93 &	19.33 &	18.42 &	6.36 & 13.56\\
    $1$/$4$-SANet + SKT & {\bf\color{red}3.42} & {\bf\color{red}16.13} & {\bf\color{red}15.82} & {\bf\color{red}15.37} & {\bf\color{red}4.91} &	{\bf\color{red}11.13} \\
    \hline
    CSRNet* & 1.58 &	13.55 &	14.70 &	7.29 & 3.28 & 8.08  \\
    $1$/$4$-CSRNet   & 1.96 &	15.70 &	20.59 &	8.52 & 3.70 & 10.09  \\
    $1$/$4$-CSRNet + SKT           & {\bf\color{red}1.77} & {\bf\color{red}12.32} & {\bf\color{red}14.49} & {\bf\color{red}7.87} & {\bf\color{red}3.10} & {\bf\color{red}7.91} \\
    \hline
    BL* & 1.79	& 10.70 & 14.12 & 	7.08 & 	3.19 &	7.37  \\
    $1$/$4$-BL & 1.97 &	18.39 &	28.95 &	8.12 &	3.94 &	12.27   \\
    $1$/$4$-BL + SKT & {\bf\color{red}1.41} &	{\bf\color{red}10.45} &	{\bf\color{red}13.10} &	{\bf\color{red}7.63} &	{\bf\color{red}4.08} &	{\bf\color{red}7.34} \\
    \hline
    \end{tabular}
  }
  \vspace{1mm}
  \caption{MAE of different methods on the WorldExpo'10 dataset. The models with symbol * are our reimplemented teacher networks.}
  \vspace{-3mm}
  \label{tab:WorldExpo}
\end{table}

\subsubsection{\bf{Performance Comparison}}
The performance comparison with recent state-of-the-art methods on Shanghaitech, UCF-QNRF and WorldExpo'10 datasets are reported in Tables \ref{tab:ShanghaiTech}, \ref{tab:UCF-QNRF} and \ref{tab:WorldExpo}, respectively.
As can be observed, the BL model is the existing best-performing method, achieving the lowest MAE and RMSE on almost all these datasets. The CSRNet and SANet also show a relatively good performance among the compared methods.
However, when reduced in model size to gain efficiency, the 1/4-BL, 1/4-CSRNet and 1/4-SANet models without knowledge transfer have a heavy performance degradation, compared with the original models.
By applying our SKT method, these lightweight models can obtain comparable results with the original models, and even achieve better performance on some datasets.
For example, as shown in Table \ref{tab:ShanghaiTech}, our 1/4-CSRNet+SKT performs better in both MAE and RMSE than the original CSRNet on Shanghaitech Part-B, while 1/4-BL+SKT obtains a new state-of-the-art RMSE of 102.33 on Shanghaitech Part-A.
It can also be observed from Table \ref{tab:UCF-QNRF} that 1/4-BL+SKT achieves an impressive state-of-the-art RMSE of 156.82 on the UCF-QNRF dataset.
Such superior performance is attributed to following reasons: {\bf{i)}} Our distilled models fully absorbs the structured knowledge of the teacher networks; {\bf{ii)}} Having less parameters, these models can effectively alleviate the overfitting of crowd counting.
%

\subsubsection{\bf{Efficiency Comparison}}

A critical goal of this work is to achieve model efficiency. To further verify the superiority of SKT. we also compare our method with existing crowd counting models on inference efficiency. In Table \ref{tab:efficiency-all}, we summarize the model sizes and the inference efficiencies of different models. Specifically, the inference time of using GPU or only CPU to process an image with the average resolution for each dataset are reported comprehensively, along with the number of the consumed FLOPs. The average resolution of images on each dataset is listed in the first row of Table \ref{tab:efficiency-all}.

As can be observed, all original models except SANet have a large number of parameters. When we compress these models with the proposed SKT, the generated models have a 16$\times$ reduction in model size and FLOPs, meanwhile achieving an order of magnitude speed-up.
For example, when testing a 2032$\times$2912 image from UCF-QNRF, our 1/4-BL+SKT only requires 90.96 milliseconds on GPU and 9.78 seconds on CPU, being 6.5/13.4$\times$ faster than the original BL model. On Shanghaitech Part-A, our 1/4-CSRNet+SKT takes 13.09 milliseconds on GPU (7.5$\times$ speed-up) and 8.88 seconds on CPU (9.0$\times$ speed-up) to process a 576$\times$864 image.
Interestingly, we find that 1/4-SANet+SKT runs slower than 1/4-BL+SKT, although SANet is much faster than BL. It mainly results from that 1/4-SANet+SKT has many stacked/parallel features with small volumes, and feature communication/synchronization consumes some extra time.
In summary, the distilled VGG-based models can achieve very impressive efficiencies and satisfactory performance.

\section{Conclusion}

In this work, we propose a general Structured Knowledge Transfer (SKT) framework to improve the efficiencies of existing crowd counting models. Specifically, an Intra-Layer Pattern Transfer and an Inter-Layer Relation Transfer are incorporated to fully transfer the structured knowledge (e.g., layer-wise one and and cross-layer one) from a heavy teacher network to a lightweight student network.
Extensive evaluations on three standard benchmarks show that the proposed SKT can efficiently compress extensive models of crowd counting (e.g, CSRNet, BL and SANet).
In particular, our distilled VGG-based models can achieve at least 6.5$\times$ speed-up on GPU and 9.0$\times$ speed-up on CPU, and meanwhile preserve very competitive performance.

\begin{acks}
This work was supported in part by National Natural Science Foundation of China (NSFC) under Grant No. U1811463, 61876045 and 61976250, in part by the Natural Science Foundation of Guangdong Province under Grant No. 2017A030312006, in part by the Guangdong Basic and Applied Basic Research Foundation under Grant No. 2020B1515020048, and in part by Zhujiang Science and Technology New Star Project of Guangzhou under Grant No. 201906010057.
\end{acks}

\bibliographystyle{ACM-Reference-Format}
\bibliography{acmart}


\begin{thebibliography}{78}


\ifx \showCODEN    \undefined \def \showCODEN     #1{\unskip}     \fi
\ifx \showDOI      \undefined \def \showDOI       #1{#1}\fi
\ifx \showISBNx    \undefined \def \showISBNx     #1{\unskip}     \fi
\ifx \showISBNxiii \undefined \def \showISBNxiii  #1{\unskip}     \fi
\ifx \showISSN     \undefined \def \showISSN      #1{\unskip}     \fi
\ifx \showLCCN     \undefined \def \showLCCN      #1{\unskip}     \fi
\ifx \shownote     \undefined \def \shownote      #1{#1}          \fi
\ifx \showarticletitle \undefined \def \showarticletitle #1{#1}   \fi
\ifx \showURL      \undefined \def \showURL       {\relax}        \fi
\providecommand\bibfield[2]{#2}
\providecommand\bibinfo[2]{#2}
\providecommand\natexlab[1]{#1}
\providecommand\showeprint[2][]{arXiv:#2}

\bibitem[\protect\citeauthoryear{??}{IOT}{[n.d.]}]%
        {IOT}
 \bibinfo{year}{[n.d.]}\natexlab{}.
\newblock
  \bibinfo{howpublished}{\url{https://en.wikipedia.org/wiki/Edge_computing}}.
\newblock


\bibitem[\protect\citeauthoryear{Babu~Sam, Sajjan, Venkatesh~Babu, and
  Srinivasan}{Babu~Sam et~al\mbox{.}}{2018}]%
        {babu2018divide}
\bibfield{author}{\bibinfo{person}{Deepak Babu~Sam}, \bibinfo{person}{Neeraj~N
  Sajjan}, \bibinfo{person}{R Venkatesh~Babu}, {and} \bibinfo{person}{Mukundhan
  Srinivasan}.} \bibinfo{year}{2018}\natexlab{}.
\newblock \showarticletitle{Divide and Grow: Capturing Huge Diversity in Crowd
  Images With Incrementally Growing CNN}. In \bibinfo{booktitle}{\emph{CVPR}}.
  \bibinfo{pages}{3618--3626}.
\newblock


\bibitem[\protect\citeauthoryear{Badrinarayanan, Kendall, and
  Cipolla}{Badrinarayanan et~al\mbox{.}}{2015}]%
        {badrinarayanan2015segnet}
\bibfield{author}{\bibinfo{person}{Vijay Badrinarayanan}, \bibinfo{person}{Alex
  Kendall}, {and} \bibinfo{person}{Roberto Cipolla}.}
  \bibinfo{year}{2015}\natexlab{}.
\newblock \showarticletitle{Segnet: A deep convolutional encoder-decoder
  architecture for image segmentation}.
\newblock \bibinfo{journal}{\emph{arXiv preprint arXiv:1511.00561}}
  (\bibinfo{year}{2015}).
\newblock


\bibitem[\protect\citeauthoryear{Boominathan, Kruthiventi, and
  Babu}{Boominathan et~al\mbox{.}}{2016}]%
        {boominathan2016crowdnet}
\bibfield{author}{\bibinfo{person}{Lokesh Boominathan},
  \bibinfo{person}{Srinivas~SS Kruthiventi}, {and} \bibinfo{person}{R~Venkatesh
  Babu}.} \bibinfo{year}{2016}\natexlab{}.
\newblock \showarticletitle{Crowdnet: A deep convolutional network for dense
  crowd counting}. In \bibinfo{booktitle}{\emph{ACM MM}}. ACM,
  \bibinfo{pages}{640--644}.
\newblock


\bibitem[\protect\citeauthoryear{Cai, He, Sun, and Vasconcelos}{Cai
  et~al\mbox{.}}{2017}]%
        {cai2017deep}
\bibfield{author}{\bibinfo{person}{Zhaowei Cai}, \bibinfo{person}{Xiaodong He},
  \bibinfo{person}{Jian Sun}, {and} \bibinfo{person}{Nuno Vasconcelos}.}
  \bibinfo{year}{2017}\natexlab{}.
\newblock \showarticletitle{Deep learning with low precision by half-wave
  gaussian quantization}. In \bibinfo{booktitle}{\emph{CVPR}}.
  \bibinfo{pages}{5918--5926}.
\newblock


\bibitem[\protect\citeauthoryear{Cao, Wang, Zhao, and Su}{Cao
  et~al\mbox{.}}{2018}]%
        {cao2018scale}
\bibfield{author}{\bibinfo{person}{Xinkun Cao}, \bibinfo{person}{Zhipeng Wang},
  \bibinfo{person}{Yanyun Zhao}, {and} \bibinfo{person}{Fei Su}.}
  \bibinfo{year}{2018}\natexlab{}.
\newblock \showarticletitle{Scale Aggregation Network for Accurate and
  Efficient Crowd Counting}. In \bibinfo{booktitle}{\emph{ECCV}}.
  \bibinfo{pages}{734--750}.
\newblock


\bibitem[\protect\citeauthoryear{Chen, Loy, Gong, and Xiang}{Chen
  et~al\mbox{.}}{2012}]%
        {chen2012feature}
\bibfield{author}{\bibinfo{person}{Ke Chen}, \bibinfo{person}{Chen~Change Loy},
  \bibinfo{person}{Shaogang Gong}, {and} \bibinfo{person}{Tony Xiang}.}
  \bibinfo{year}{2012}\natexlab{}.
\newblock \showarticletitle{Feature Mining for Localised Crowd Counting.}. In
  \bibinfo{booktitle}{\emph{BMVC}}, Vol.~\bibinfo{volume}{1}.
  \bibinfo{pages}{3}.
\newblock


\bibitem[\protect\citeauthoryear{Chen, Lin, Zuo, Luo, and Zhang}{Chen
  et~al\mbox{.}}{2018a}]%
        {chen2018learning}
\bibfield{author}{\bibinfo{person}{Tianshui Chen}, \bibinfo{person}{Liang Lin},
  \bibinfo{person}{Wangmeng Zuo}, \bibinfo{person}{Xiaonan Luo}, {and}
  \bibinfo{person}{Lei Zhang}.} \bibinfo{year}{2018}\natexlab{a}.
\newblock \showarticletitle{Learning a wavelet-like auto-encoder to accelerate
  deep neural networks}. In \bibinfo{booktitle}{\emph{AAAI}}.
\newblock


\bibitem[\protect\citeauthoryear{Chen, Wu, Gao, Dong, Luo, and Lin}{Chen
  et~al\mbox{.}}{2018b}]%
        {chen2018fine}
\bibfield{author}{\bibinfo{person}{Tianshui Chen}, \bibinfo{person}{Wenxi Wu},
  \bibinfo{person}{Yuefang Gao}, \bibinfo{person}{Le Dong},
  \bibinfo{person}{Xiaonan Luo}, {and} \bibinfo{person}{Liang Lin}.}
  \bibinfo{year}{2018}\natexlab{b}.
\newblock \showarticletitle{Fine-grained representation learning and
  recognition by exploiting hierarchical semantic embedding}. In
  \bibinfo{booktitle}{\emph{Proceedings of the 26th ACM international
  conference on Multimedia}}. \bibinfo{pages}{2023--2031}.
\newblock


\bibitem[\protect\citeauthoryear{Cheng, Li, Dai, Wu, He, and Hauptmann}{Cheng
  et~al\mbox{.}}{2019}]%
        {cheng2019improving}
\bibfield{author}{\bibinfo{person}{Zhi-Qi Cheng}, \bibinfo{person}{Jun-Xiu Li},
  \bibinfo{person}{Qi Dai}, \bibinfo{person}{Xiao Wu}, \bibinfo{person}{Jun-Yan
  He}, {and} \bibinfo{person}{Alexander~G Hauptmann}.}
  \bibinfo{year}{2019}\natexlab{}.
\newblock \showarticletitle{Improving the learning of multi-column
  convolutional neural network for crowd counting}. In
  \bibinfo{booktitle}{\emph{Proceedings of the 27th ACM International
  Conference on Multimedia}}. \bibinfo{pages}{1897--1906}.
\newblock


\bibitem[\protect\citeauthoryear{Ge and Collins}{Ge and Collins}{2009}]%
        {ge2009marked}
\bibfield{author}{\bibinfo{person}{Weina Ge} {and} \bibinfo{person}{Robert~T
  Collins}.} \bibinfo{year}{2009}\natexlab{}.
\newblock \showarticletitle{Marked point processes for crowd counting}. In
  \bibinfo{booktitle}{\emph{CVPR}}. IEEE, \bibinfo{pages}{2913--2920}.
\newblock


\bibitem[\protect\citeauthoryear{Gong, Liu, Yang, and Bourdev}{Gong
  et~al\mbox{.}}{2014}]%
        {gong2014compressing}
\bibfield{author}{\bibinfo{person}{Yunchao Gong}, \bibinfo{person}{Liu Liu},
  \bibinfo{person}{Ming Yang}, {and} \bibinfo{person}{Lubomir Bourdev}.}
  \bibinfo{year}{2014}\natexlab{}.
\newblock \showarticletitle{Compressing deep convolutional networks using
  vector quantization}.
\newblock \bibinfo{journal}{\emph{arXiv preprint arXiv:1412.6115}}
  (\bibinfo{year}{2014}).
\newblock


\bibitem[\protect\citeauthoryear{Guo, Li, Zha, and Wang}{Guo
  et~al\mbox{.}}{2019}]%
        {guo2019dadnet}
\bibfield{author}{\bibinfo{person}{Dan Guo}, \bibinfo{person}{Kun Li},
  \bibinfo{person}{Zheng-Jun Zha}, {and} \bibinfo{person}{Meng Wang}.}
  \bibinfo{year}{2019}\natexlab{}.
\newblock \showarticletitle{Dadnet: Dilated-attention-deformable convnet for
  crowd counting}. In \bibinfo{booktitle}{\emph{Proceedings of the 27th ACM
  International Conference on Multimedia}}. \bibinfo{pages}{1823--1832}.
\newblock


\bibitem[\protect\citeauthoryear{Han, Mao, and Dally}{Han
  et~al\mbox{.}}{2015}]%
        {han2015deep}
\bibfield{author}{\bibinfo{person}{Song Han}, \bibinfo{person}{Huizi Mao},
  {and} \bibinfo{person}{William~J Dally}.} \bibinfo{year}{2015}\natexlab{}.
\newblock \showarticletitle{Deep compression: Compressing deep neural networks
  with pruning, trained quantization and huffman coding}.
\newblock \bibinfo{journal}{\emph{arXiv preprint arXiv:1510.00149}}
  (\bibinfo{year}{2015}).
\newblock


\bibitem[\protect\citeauthoryear{He, Zhang, Ren, and Sun}{He
  et~al\mbox{.}}{2016}]%
        {he2016deep}
\bibfield{author}{\bibinfo{person}{Kaiming He}, \bibinfo{person}{Xiangyu
  Zhang}, \bibinfo{person}{Shaoqing Ren}, {and} \bibinfo{person}{Jian Sun}.}
  \bibinfo{year}{2016}\natexlab{}.
\newblock \showarticletitle{Deep residual learning for image recognition}. In
  \bibinfo{booktitle}{\emph{CVPR}}. \bibinfo{pages}{770--778}.
\newblock


\bibitem[\protect\citeauthoryear{He, Shen, Tian, Gong, Sun, and Yan}{He
  et~al\mbox{.}}{2019}]%
        {he2019knowledge}
\bibfield{author}{\bibinfo{person}{Tong He}, \bibinfo{person}{Chunhua Shen},
  \bibinfo{person}{Zhi Tian}, \bibinfo{person}{Dong Gong},
  \bibinfo{person}{Changming Sun}, {and} \bibinfo{person}{Youliang Yan}.}
  \bibinfo{year}{2019}\natexlab{}.
\newblock \showarticletitle{Knowledge adaptation for efficient semantic
  segmentation}. In \bibinfo{booktitle}{\emph{Proceedings of the IEEE
  Conference on Computer Vision and Pattern Recognition}}.
  \bibinfo{pages}{578--587}.
\newblock


\bibitem[\protect\citeauthoryear{He, Zhang, and Sun}{He et~al\mbox{.}}{2017}]%
        {he2017channel}
\bibfield{author}{\bibinfo{person}{Yihui He}, \bibinfo{person}{Xiangyu Zhang},
  {and} \bibinfo{person}{Jian Sun}.} \bibinfo{year}{2017}\natexlab{}.
\newblock \showarticletitle{Channel pruning for accelerating very deep neural
  networks}. In \bibinfo{booktitle}{\emph{ICCV}}. \bibinfo{pages}{1389--1397}.
\newblock


\bibitem[\protect\citeauthoryear{Heo, Lee, Yun, and Choi}{Heo
  et~al\mbox{.}}{2019}]%
        {heo2019knowledge}
\bibfield{author}{\bibinfo{person}{Byeongho Heo}, \bibinfo{person}{Minsik Lee},
  \bibinfo{person}{Sangdoo Yun}, {and} \bibinfo{person}{Jin~Young Choi}.}
  \bibinfo{year}{2019}\natexlab{}.
\newblock \showarticletitle{Knowledge transfer via distillation of activation
  boundaries formed by hidden neurons}. In \bibinfo{booktitle}{\emph{AAAI}},
  Vol.~\bibinfo{volume}{33}. \bibinfo{pages}{3779--3787}.
\newblock


\bibitem[\protect\citeauthoryear{Hinton, Vinyals, and Dean}{Hinton
  et~al\mbox{.}}{2015}]%
        {hinton2015distilling}
\bibfield{author}{\bibinfo{person}{Geoffrey~E Hinton}, \bibinfo{person}{Oriol
  Vinyals}, {and} \bibinfo{person}{Jeffrey Dean}.}
  \bibinfo{year}{2015}\natexlab{}.
\newblock \showarticletitle{Distilling the Knowledge in a Neural Network}.
\newblock \bibinfo{journal}{\emph{arXiv: Machine Learning}}
  (\bibinfo{year}{2015}).
\newblock


\bibitem[\protect\citeauthoryear{Huang and Wang}{Huang and Wang}{2017}]%
        {huang2017like}
\bibfield{author}{\bibinfo{person}{Zehao Huang} {and} \bibinfo{person}{Naiyan
  Wang}.} \bibinfo{year}{2017}\natexlab{}.
\newblock \showarticletitle{Like what you like: Knowledge distill via neuron
  selectivity transfer}.
\newblock \bibinfo{journal}{\emph{arXiv preprint arXiv:1707.01219}}
  (\bibinfo{year}{2017}).
\newblock


\bibitem[\protect\citeauthoryear{Idrees, Saleemi, Seibert, and Shah}{Idrees
  et~al\mbox{.}}{2013}]%
        {idrees2013multi}
\bibfield{author}{\bibinfo{person}{Haroon Idrees}, \bibinfo{person}{Imran
  Saleemi}, \bibinfo{person}{Cody Seibert}, {and} \bibinfo{person}{Mubarak
  Shah}.} \bibinfo{year}{2013}\natexlab{}.
\newblock \showarticletitle{Multi-source multi-scale counting in extremely
  dense crowd images}. In \bibinfo{booktitle}{\emph{CVPR}}.
  \bibinfo{pages}{2547--2554}.
\newblock


\bibitem[\protect\citeauthoryear{Idrees, Tayyab, Athrey, Zhang, Al-Maadeed,
  Rajpoot, and Shah}{Idrees et~al\mbox{.}}{2018}]%
        {idrees2018composition}
\bibfield{author}{\bibinfo{person}{Haroon Idrees}, \bibinfo{person}{Muhmmad
  Tayyab}, \bibinfo{person}{Kishan Athrey}, \bibinfo{person}{Dong Zhang},
  \bibinfo{person}{Somaya Al-Maadeed}, \bibinfo{person}{Nasir Rajpoot}, {and}
  \bibinfo{person}{Mubarak Shah}.} \bibinfo{year}{2018}\natexlab{}.
\newblock \showarticletitle{Composition Loss for Counting, Density Map
  Estimation and Localization in Dense Crowds}. In
  \bibinfo{booktitle}{\emph{ECCV}}.
\newblock


\bibitem[\protect\citeauthoryear{Jacob, Kligys, Chen, Zhu, Tang, Howard, Adam,
  and Kalenichenko}{Jacob et~al\mbox{.}}{2018}]%
        {jacob2018quantization}
\bibfield{author}{\bibinfo{person}{Benoit Jacob}, \bibinfo{person}{Skirmantas
  Kligys}, \bibinfo{person}{Bo Chen}, \bibinfo{person}{Menglong Zhu},
  \bibinfo{person}{Matthew Tang}, \bibinfo{person}{Andrew Howard},
  \bibinfo{person}{Hartwig Adam}, {and} \bibinfo{person}{Dmitry Kalenichenko}.}
  \bibinfo{year}{2018}\natexlab{}.
\newblock \showarticletitle{Quantization and training of neural networks for
  efficient integer-arithmetic-only inference}. In
  \bibinfo{booktitle}{\emph{CVPR}}. \bibinfo{pages}{2704--2713}.
\newblock


\bibitem[\protect\citeauthoryear{Jiang, Xiao, Zhang, Zhen, Cao, Doermann, and
  Shao}{Jiang et~al\mbox{.}}{2019}]%
        {jiang2019crowd}
\bibfield{author}{\bibinfo{person}{Xiaolong Jiang}, \bibinfo{person}{Zehao
  Xiao}, \bibinfo{person}{Baochang Zhang}, \bibinfo{person}{Xiantong Zhen},
  \bibinfo{person}{Xianbin Cao}, \bibinfo{person}{David Doermann}, {and}
  \bibinfo{person}{Ling Shao}.} \bibinfo{year}{2019}\natexlab{}.
\newblock \showarticletitle{Crowd Counting and Density Estimation by Trellis
  Encoder-Decoder Networks}. In \bibinfo{booktitle}{\emph{CVPR}}.
  \bibinfo{pages}{6133--6142}.
\newblock


\bibitem[\protect\citeauthoryear{Li, Kadav, Durdanovic, Samet, and Graf}{Li
  et~al\mbox{.}}{2016}]%
        {li2016pruning}
\bibfield{author}{\bibinfo{person}{Hao Li}, \bibinfo{person}{Asim Kadav},
  \bibinfo{person}{Igor Durdanovic}, \bibinfo{person}{Hanan Samet}, {and}
  \bibinfo{person}{Hans~Peter Graf}.} \bibinfo{year}{2016}\natexlab{}.
\newblock \showarticletitle{Pruning filters for efficient convnets}.
\newblock \bibinfo{journal}{\emph{arXiv preprint arXiv:1608.08710}}
  (\bibinfo{year}{2016}).
\newblock


\bibitem[\protect\citeauthoryear{Li, Zhang, Huang, and Tan}{Li
  et~al\mbox{.}}{2008}]%
        {li2008estimating}
\bibfield{author}{\bibinfo{person}{Min Li}, \bibinfo{person}{Zhaoxiang Zhang},
  \bibinfo{person}{Kaiqi Huang}, {and} \bibinfo{person}{Tieniu Tan}.}
  \bibinfo{year}{2008}\natexlab{}.
\newblock \showarticletitle{Estimating the number of people in crowded scenes
  by mid based foreground segmentation and head-shoulder detection}. In
  \bibinfo{booktitle}{\emph{ICPR}}. IEEE, \bibinfo{pages}{1--4}.
\newblock


\bibitem[\protect\citeauthoryear{Li, Chang, Wang, Ni, Hong, and Yan}{Li
  et~al\mbox{.}}{2015}]%
        {li2015crowded}
\bibfield{author}{\bibinfo{person}{Teng Li}, \bibinfo{person}{Huan Chang},
  \bibinfo{person}{Meng Wang}, \bibinfo{person}{Bingbing Ni},
  \bibinfo{person}{Richang Hong}, {and} \bibinfo{person}{Shuicheng Yan}.}
  \bibinfo{year}{2015}\natexlab{}.
\newblock \showarticletitle{Crowded Scene Analysis: A Survey}.
\newblock \bibinfo{journal}{\emph{T-CSVT}} \bibinfo{volume}{25},
  \bibinfo{number}{3} (\bibinfo{year}{2015}), \bibinfo{pages}{367--386}.
\newblock


\bibitem[\protect\citeauthoryear{Li, Zhang, and Chen}{Li et~al\mbox{.}}{2018}]%
        {li2018csrnet}
\bibfield{author}{\bibinfo{person}{Yuhong Li}, \bibinfo{person}{Xiaofan Zhang},
  {and} \bibinfo{person}{Deming Chen}.} \bibinfo{year}{2018}\natexlab{}.
\newblock \showarticletitle{CSRNet: Dilated convolutional neural networks for
  understanding the highly congested scenes}. In
  \bibinfo{booktitle}{\emph{CVPR}}. \bibinfo{pages}{1091--1100}.
\newblock


\bibitem[\protect\citeauthoryear{Lian, Li, Zheng, Luo, and Gao}{Lian
  et~al\mbox{.}}{2019}]%
        {lian2019density}
\bibfield{author}{\bibinfo{person}{Dongze Lian}, \bibinfo{person}{Jing Li},
  \bibinfo{person}{Jia Zheng}, \bibinfo{person}{Weixin Luo}, {and}
  \bibinfo{person}{Shenghua Gao}.} \bibinfo{year}{2019}\natexlab{}.
\newblock \showarticletitle{Density Map Regression Guided Detection Network for
  RGB-D Crowd Counting and Localization}. In \bibinfo{booktitle}{\emph{CVPR}}.
  \bibinfo{pages}{1821--1830}.
\newblock


\bibitem[\protect\citeauthoryear{Liu, Gao, Meng, and Hauptmann}{Liu
  et~al\mbox{.}}{2018a}]%
        {liu2018decidenet}
\bibfield{author}{\bibinfo{person}{Jiang Liu}, \bibinfo{person}{Chenqiang Gao},
  \bibinfo{person}{Deyu Meng}, {and} \bibinfo{person}{Alexander~G Hauptmann}.}
  \bibinfo{year}{2018}\natexlab{a}.
\newblock \showarticletitle{Decidenet: Counting varying density crowds through
  attention guided detection and density estimation}. In
  \bibinfo{booktitle}{\emph{CVPR}}. \bibinfo{pages}{5197--5206}.
\newblock


\bibitem[\protect\citeauthoryear{Liu, Qiu, Li, Liu, Ouyang, and Lin}{Liu
  et~al\mbox{.}}{2019c}]%
        {liu2019crowd}
\bibfield{author}{\bibinfo{person}{Lingbo Liu}, \bibinfo{person}{Zhilin Qiu},
  \bibinfo{person}{Guanbin Li}, \bibinfo{person}{Shufan Liu},
  \bibinfo{person}{Wanli Ouyang}, {and} \bibinfo{person}{Liang Lin}.}
  \bibinfo{year}{2019}\natexlab{c}.
\newblock \showarticletitle{Crowd Counting with Deep Structured Scale
  Integration Network}. In \bibinfo{booktitle}{\emph{ICCV}}.
  \bibinfo{pages}{1774--1783}.
\newblock


\bibitem[\protect\citeauthoryear{Liu, Wang, Li, Ouyang, and Lin}{Liu
  et~al\mbox{.}}{2018d}]%
        {liu2018crowd}
\bibfield{author}{\bibinfo{person}{Lingbo Liu}, \bibinfo{person}{Hongjun Wang},
  \bibinfo{person}{Guanbin Li}, \bibinfo{person}{Wanli Ouyang}, {and}
  \bibinfo{person}{Liang Lin}.} \bibinfo{year}{2018}\natexlab{d}.
\newblock \showarticletitle{Crowd counting using deep recurrent spatial-aware
  network}. In \bibinfo{booktitle}{\emph{IJCAI}}.
\newblock


\bibitem[\protect\citeauthoryear{Liu, Zhang, Peng, Li, Du, and Lin}{Liu
  et~al\mbox{.}}{2018e}]%
        {liu2018attentive}
\bibfield{author}{\bibinfo{person}{Lingbo Liu}, \bibinfo{person}{Ruimao Zhang},
  \bibinfo{person}{Jiefeng Peng}, \bibinfo{person}{Guanbin Li},
  \bibinfo{person}{Bowen Du}, {and} \bibinfo{person}{Liang Lin}.}
  \bibinfo{year}{2018}\natexlab{e}.
\newblock \showarticletitle{Attentive Crowd Flow Machines}. In
  \bibinfo{booktitle}{\emph{ACM MM}}. ACM, \bibinfo{pages}{1553--1561}.
\newblock


\bibitem[\protect\citeauthoryear{Liu, Zhen, Li, Zhan, He, Du, and Lin}{Liu
  et~al\mbox{.}}{2020}]%
        {liu2020dynamic}
\bibfield{author}{\bibinfo{person}{Lingbo Liu}, \bibinfo{person}{Jiajie Zhen},
  \bibinfo{person}{Guanbin Li}, \bibinfo{person}{Geng Zhan},
  \bibinfo{person}{Zhaocheng He}, \bibinfo{person}{Bowen Du}, {and}
  \bibinfo{person}{Liang Lin}.} \bibinfo{year}{2020}\natexlab{}.
\newblock \showarticletitle{Dynamic Spatial-Temporal Representation Learning
  for Traffic Flow Prediction}.
\newblock \bibinfo{journal}{\emph{IEEE Transactions on Intelligent
  Transportation Systems}} (\bibinfo{year}{2020}).
\newblock


\bibitem[\protect\citeauthoryear{Liu, Long, Zou, Niu, Pan, and Wu}{Liu
  et~al\mbox{.}}{2019b}]%
        {liu2019adcrowdnet}
\bibfield{author}{\bibinfo{person}{Ning Liu}, \bibinfo{person}{Yongchao Long},
  \bibinfo{person}{Changqing Zou}, \bibinfo{person}{Qun Niu},
  \bibinfo{person}{Li Pan}, {and} \bibinfo{person}{Hefeng Wu}.}
  \bibinfo{year}{2019}\natexlab{b}.
\newblock \showarticletitle{ADCrowdNet: An Attention-injective Deformable
  Convolutional Network for Crowd Understanding}. In
  \bibinfo{booktitle}{\emph{CVPR}}. \bibinfo{pages}{3225--3234}.
\newblock


\bibitem[\protect\citeauthoryear{Liu, Salzmann, and Fua}{Liu
  et~al\mbox{.}}{2019d}]%
        {liu2019context}
\bibfield{author}{\bibinfo{person}{Weizhe Liu}, \bibinfo{person}{Mathieu
  Salzmann}, {and} \bibinfo{person}{Pascal Fua}.}
  \bibinfo{year}{2019}\natexlab{d}.
\newblock \showarticletitle{Context-Aware Crowd Counting}. In
  \bibinfo{booktitle}{\emph{CVPR}}. \bibinfo{pages}{5099--5108}.
\newblock


\bibitem[\protect\citeauthoryear{Liu, Pool, Han, and Dally}{Liu
  et~al\mbox{.}}{2018b}]%
        {liu2018efficient}
\bibfield{author}{\bibinfo{person}{Xingyu Liu}, \bibinfo{person}{Jeff Pool},
  \bibinfo{person}{Song Han}, {and} \bibinfo{person}{William~J Dally}.}
  \bibinfo{year}{2018}\natexlab{b}.
\newblock \showarticletitle{Efficient sparse-winograd convolutional neural
  networks}.
\newblock \bibinfo{journal}{\emph{arXiv preprint arXiv:1802.06367}}
  (\bibinfo{year}{2018}).
\newblock


\bibitem[\protect\citeauthoryear{Liu, van~de Weijer, and Bagdanov}{Liu
  et~al\mbox{.}}{2018c}]%
        {liu2018leveraging}
\bibfield{author}{\bibinfo{person}{Xialei Liu}, \bibinfo{person}{Joost van~de
  Weijer}, {and} \bibinfo{person}{Andrew~D Bagdanov}.}
  \bibinfo{year}{2018}\natexlab{c}.
\newblock \showarticletitle{Leveraging Unlabeled Data for Crowd Counting by
  Learning to Rank}. In \bibinfo{booktitle}{\emph{CVPR}}.
\newblock


\bibitem[\protect\citeauthoryear{Liu, Chen, Liu, Qin, Luo, and Wang}{Liu
  et~al\mbox{.}}{2019a}]%
        {liu2019structured}
\bibfield{author}{\bibinfo{person}{Yifan Liu}, \bibinfo{person}{Ke Chen},
  \bibinfo{person}{Chris Liu}, \bibinfo{person}{Zengchang Qin},
  \bibinfo{person}{Zhenbo Luo}, {and} \bibinfo{person}{Jingdong Wang}.}
  \bibinfo{year}{2019}\natexlab{a}.
\newblock \showarticletitle{Structured knowledge distillation for semantic
  segmentation}. In \bibinfo{booktitle}{\emph{Proceedings of the IEEE
  Conference on Computer Vision and Pattern Recognition}}.
  \bibinfo{pages}{2604--2613}.
\newblock


\bibitem[\protect\citeauthoryear{Ma, Wei, Hong, and Gong}{Ma
  et~al\mbox{.}}{2019}]%
        {ma2019bayesian}
\bibfield{author}{\bibinfo{person}{Zhiheng Ma}, \bibinfo{person}{Xing Wei},
  \bibinfo{person}{Xiaopeng Hong}, {and} \bibinfo{person}{Yihong Gong}.}
  \bibinfo{year}{2019}\natexlab{}.
\newblock \showarticletitle{Bayesian loss for crowd count estimation with point
  supervision}. In \bibinfo{booktitle}{\emph{ICCV}}.
  \bibinfo{pages}{6142--6151}.
\newblock


\bibitem[\protect\citeauthoryear{Ma, Wei, Hong, and Gong}{Ma
  et~al\mbox{.}}{2020}]%
        {ma2020learning}
\bibfield{author}{\bibinfo{person}{Zhiheng Ma}, \bibinfo{person}{Xing Wei},
  \bibinfo{person}{Xiaopeng Hong}, {and} \bibinfo{person}{Yihong Gong}.}
  \bibinfo{year}{2020}\natexlab{}.
\newblock \showarticletitle{Learning Scales from Points: A Scale-aware
  Probabilistic Model for Crowd Counting}. In \bibinfo{booktitle}{\emph{ACM
  International Conference on Multimedia}}.
\newblock


\bibitem[\protect\citeauthoryear{Mirzadeh, Farajtabar, Li, and
  Ghasemzadeh}{Mirzadeh et~al\mbox{.}}{2019}]%
        {mirzadeh2019improved}
\bibfield{author}{\bibinfo{person}{Seyed-Iman Mirzadeh},
  \bibinfo{person}{Mehrdad Farajtabar}, \bibinfo{person}{Ang Li}, {and}
  \bibinfo{person}{Hassan Ghasemzadeh}.} \bibinfo{year}{2019}\natexlab{}.
\newblock \showarticletitle{Improved knowledge distillation via teacher
  assistant: Bridging the gap between student and teacher}.
\newblock \bibinfo{journal}{\emph{arXiv preprint arXiv:1902.03393}}
  (\bibinfo{year}{2019}).
\newblock


\bibitem[\protect\citeauthoryear{Onoro-Rubio and L{\'o}pez-Sastre}{Onoro-Rubio
  and L{\'o}pez-Sastre}{2016}]%
        {onoro2016towards}
\bibfield{author}{\bibinfo{person}{Daniel Onoro-Rubio} {and}
  \bibinfo{person}{Roberto~J L{\'o}pez-Sastre}.}
  \bibinfo{year}{2016}\natexlab{}.
\newblock \showarticletitle{Towards perspective-free object counting with deep
  learning}. In \bibinfo{booktitle}{\emph{ECCV}}. Springer,
  \bibinfo{pages}{615--629}.
\newblock


\bibitem[\protect\citeauthoryear{Qiu, Liu, Li, Wang, Xiao, and Lin}{Qiu
  et~al\mbox{.}}{2019}]%
        {qiu2019crowd}
\bibfield{author}{\bibinfo{person}{Zhilin Qiu}, \bibinfo{person}{Lingbo Liu},
  \bibinfo{person}{Guanbin Li}, \bibinfo{person}{Qing Wang},
  \bibinfo{person}{Nong Xiao}, {and} \bibinfo{person}{Liang Lin}.}
  \bibinfo{year}{2019}\natexlab{}.
\newblock \showarticletitle{Crowd counting via multi-view scale aggregation
  networks}. In \bibinfo{booktitle}{\emph{ICME}}. IEEE,
  \bibinfo{pages}{1498--1503}.
\newblock


\bibitem[\protect\citeauthoryear{Ranjan, Le, and Hoai}{Ranjan
  et~al\mbox{.}}{2018}]%
        {ranjan2018iterative}
\bibfield{author}{\bibinfo{person}{Viresh Ranjan}, \bibinfo{person}{Hieu Le},
  {and} \bibinfo{person}{Minh Hoai}.} \bibinfo{year}{2018}\natexlab{}.
\newblock \showarticletitle{Iterative Crowd Counting}. In
  \bibinfo{booktitle}{\emph{ECCV}}.
\newblock


\bibitem[\protect\citeauthoryear{Romero, Ballas, Kahou, Chassang, Gatta, and
  Bengio}{Romero et~al\mbox{.}}{2014}]%
        {romero2014fitnets}
\bibfield{author}{\bibinfo{person}{Adriana Romero}, \bibinfo{person}{Nicolas
  Ballas}, \bibinfo{person}{Samira~Ebrahimi Kahou}, \bibinfo{person}{Antoine
  Chassang}, \bibinfo{person}{Carlo Gatta}, {and} \bibinfo{person}{Yoshua
  Bengio}.} \bibinfo{year}{2014}\natexlab{}.
\newblock \showarticletitle{Fitnets: Hints for thin deep nets}.
\newblock \bibinfo{journal}{\emph{arXiv preprint arXiv:1412.6550}}
  (\bibinfo{year}{2014}).
\newblock


\bibitem[\protect\citeauthoryear{Ryan, Denman, Fookes, and Sridharan}{Ryan
  et~al\mbox{.}}{2009}]%
        {Ryan2009multiple}
\bibfield{author}{\bibinfo{person}{D. Ryan}, \bibinfo{person}{S. Denman},
  \bibinfo{person}{C. Fookes}, {and} \bibinfo{person}{S. Sridharan}.}
  \bibinfo{year}{2009}\natexlab{}.
\newblock \showarticletitle{Crowd Counting Using Multiple Local Features}. In
  \bibinfo{booktitle}{\emph{DICTA}}. \bibinfo{pages}{81--88}.
\newblock


\bibitem[\protect\citeauthoryear{Sam, Surya, and Babu}{Sam
  et~al\mbox{.}}{2017}]%
        {sam2017switching}
\bibfield{author}{\bibinfo{person}{Deepak~Babu Sam}, \bibinfo{person}{Shiv
  Surya}, {and} \bibinfo{person}{R~Venkatesh Babu}.}
  \bibinfo{year}{2017}\natexlab{}.
\newblock \showarticletitle{Switching convolutional neural network for crowd
  counting}. In \bibinfo{booktitle}{\emph{CVPR}}, Vol.~\bibinfo{volume}{1}.
  \bibinfo{pages}{6}.
\newblock


\bibitem[\protect\citeauthoryear{Sau and Balasubramanian}{Sau and
  Balasubramanian}{2016}]%
        {sau2016deep}
\bibfield{author}{\bibinfo{person}{Bharat~Bhusan Sau} {and}
  \bibinfo{person}{Vineeth~N Balasubramanian}.}
  \bibinfo{year}{2016}\natexlab{}.
\newblock \showarticletitle{Deep model compression: Distilling knowledge from
  noisy teachers}.
\newblock \bibinfo{journal}{\emph{arXiv preprint arXiv:1610.09650}}
  (\bibinfo{year}{2016}).
\newblock


\bibitem[\protect\citeauthoryear{Semertzidis, Dimitropoulos, Koutsia, and
  Grammalidis}{Semertzidis et~al\mbox{.}}{2010}]%
        {semertzidis2010video}
\bibfield{author}{\bibinfo{person}{T Semertzidis}, \bibinfo{person}{K
  Dimitropoulos}, \bibinfo{person}{A Koutsia}, {and} \bibinfo{person}{N
  Grammalidis}.} \bibinfo{year}{2010}\natexlab{}.
\newblock \showarticletitle{Video sensor network for real-time traffic
  monitoring and surveillance}.
\newblock \bibinfo{journal}{\emph{IET intelligent transport systems}}
  \bibinfo{volume}{4}, \bibinfo{number}{2} (\bibinfo{year}{2010}),
  \bibinfo{pages}{103--112}.
\newblock


\bibitem[\protect\citeauthoryear{Shang, Ai, and Bai}{Shang
  et~al\mbox{.}}{2016}]%
        {shang2016end}
\bibfield{author}{\bibinfo{person}{Chong Shang}, \bibinfo{person}{Haizhou Ai},
  {and} \bibinfo{person}{Bo Bai}.} \bibinfo{year}{2016}\natexlab{}.
\newblock \showarticletitle{End-to-end crowd counting via joint learning local
  and global count}. In \bibinfo{booktitle}{\emph{ICIP}}. IEEE,
  \bibinfo{pages}{1215--1219}.
\newblock


\bibitem[\protect\citeauthoryear{Shen, Xu, Ni, Wang, Hu, and Yang}{Shen
  et~al\mbox{.}}{2018}]%
        {shen2018crowd}
\bibfield{author}{\bibinfo{person}{Zan Shen}, \bibinfo{person}{Yi Xu},
  \bibinfo{person}{Bingbing Ni}, \bibinfo{person}{Minsi Wang},
  \bibinfo{person}{Jianguo Hu}, {and} \bibinfo{person}{Xiaokang Yang}.}
  \bibinfo{year}{2018}\natexlab{}.
\newblock \showarticletitle{Crowd Counting via Adversarial Cross-Scale
  Consistency Pursuit}. In \bibinfo{booktitle}{\emph{CVPR}}.
  \bibinfo{pages}{5245--5254}.
\newblock


\bibitem[\protect\citeauthoryear{Shi, Yang, Xu, and Chen}{Shi
  et~al\mbox{.}}{2019b}]%
        {shi2019revisiting}
\bibfield{author}{\bibinfo{person}{Miaojing Shi}, \bibinfo{person}{Zhaohui
  Yang}, \bibinfo{person}{Chao Xu}, {and} \bibinfo{person}{Qijun Chen}.}
  \bibinfo{year}{2019}\natexlab{b}.
\newblock \showarticletitle{Revisiting perspective information for efficient
  crowd counting}. In \bibinfo{booktitle}{\emph{CVPR}}.
  \bibinfo{pages}{7279--7288}.
\newblock


\bibitem[\protect\citeauthoryear{Shi, Mettes, and Snoek}{Shi
  et~al\mbox{.}}{2019a}]%
        {shi2019counting}
\bibfield{author}{\bibinfo{person}{Zenglin Shi}, \bibinfo{person}{Pascal
  Mettes}, {and} \bibinfo{person}{Cees~GM Snoek}.}
  \bibinfo{year}{2019}\natexlab{a}.
\newblock \showarticletitle{Counting with focus for free}. In
  \bibinfo{booktitle}{\emph{ICCV}}. \bibinfo{pages}{4200--4209}.
\newblock


\bibitem[\protect\citeauthoryear{Shi, Zhang, Liu, Cao, Ye, Cheng, and
  Zheng}{Shi et~al\mbox{.}}{2018}]%
        {shi2018crowd}
\bibfield{author}{\bibinfo{person}{Zenglin Shi}, \bibinfo{person}{Le Zhang},
  \bibinfo{person}{Yun Liu}, \bibinfo{person}{Xiaofeng Cao},
  \bibinfo{person}{Yangdong Ye}, \bibinfo{person}{Ming-Ming Cheng}, {and}
  \bibinfo{person}{Guoyan Zheng}.} \bibinfo{year}{2018}\natexlab{}.
\newblock \showarticletitle{Crowd Counting With Deep Negative Correlation
  Learning}. In \bibinfo{booktitle}{\emph{CVPR}}. \bibinfo{pages}{5382--5390}.
\newblock


\bibitem[\protect\citeauthoryear{Simonyan and Zisserman}{Simonyan and
  Zisserman}{2014}]%
        {simonyan2014very}
\bibfield{author}{\bibinfo{person}{Karen Simonyan} {and}
  \bibinfo{person}{Andrew Zisserman}.} \bibinfo{year}{2014}\natexlab{}.
\newblock \showarticletitle{Very deep convolutional networks for large-scale
  image recognition}.
\newblock \bibinfo{journal}{\emph{arXiv preprint arXiv:1409.1556}}
  (\bibinfo{year}{2014}).
\newblock


\bibitem[\protect\citeauthoryear{Sindagi and Patel}{Sindagi and Patel}{2017a}]%
        {sindagi2017cnn}
\bibfield{author}{\bibinfo{person}{Vishwanath~A Sindagi} {and}
  \bibinfo{person}{Vishal~M Patel}.} \bibinfo{year}{2017}\natexlab{a}.
\newblock \showarticletitle{Cnn-based cascaded multi-task learning of
  high-level prior and density estimation for crowd counting}. In
  \bibinfo{booktitle}{\emph{AVSS}}. IEEE, \bibinfo{pages}{1--6}.
\newblock


\bibitem[\protect\citeauthoryear{Sindagi and Patel}{Sindagi and Patel}{2017b}]%
        {sindagi2017generating}
\bibfield{author}{\bibinfo{person}{Vishwanath~A Sindagi} {and}
  \bibinfo{person}{Vishal~M Patel}.} \bibinfo{year}{2017}\natexlab{b}.
\newblock \showarticletitle{Generating high-quality crowd density maps using
  contextual pyramid cnns}. In \bibinfo{booktitle}{\emph{ICCV}}. IEEE,
  \bibinfo{pages}{1879--1888}.
\newblock


\bibitem[\protect\citeauthoryear{Sindagi and Patel}{Sindagi and Patel}{2019}]%
        {sindagi2019multi}
\bibfield{author}{\bibinfo{person}{Vishwanath~A Sindagi} {and}
  \bibinfo{person}{Vishal~M Patel}.} \bibinfo{year}{2019}\natexlab{}.
\newblock \showarticletitle{Multi-Level Bottom-Top and Top-Bottom Feature
  Fusion for Crowd Counting}. In \bibinfo{booktitle}{\emph{ICCV}}.
  \bibinfo{pages}{1002--1012}.
\newblock


\bibitem[\protect\citeauthoryear{Szegedy, Liu, Jia, Sermanet, Reed, Anguelov,
  Erhan, Vanhoucke, and Rabinovich}{Szegedy et~al\mbox{.}}{2015}]%
        {szegedy2015going}
\bibfield{author}{\bibinfo{person}{Christian Szegedy}, \bibinfo{person}{Wei
  Liu}, \bibinfo{person}{Yangqing Jia}, \bibinfo{person}{Pierre Sermanet},
  \bibinfo{person}{Scott Reed}, \bibinfo{person}{Dragomir Anguelov},
  \bibinfo{person}{Dumitru Erhan}, \bibinfo{person}{Vincent Vanhoucke}, {and}
  \bibinfo{person}{Andrew Rabinovich}.} \bibinfo{year}{2015}\natexlab{}.
\newblock \showarticletitle{Going deeper with convolutions}. In
  \bibinfo{booktitle}{\emph{CVPR}}. \bibinfo{pages}{1--9}.
\newblock


\bibitem[\protect\citeauthoryear{Tai, Xiao, Zhang, Wang, et~al\mbox{.}}{Tai
  et~al\mbox{.}}{2015}]%
        {tai2015convolutional}
\bibfield{author}{\bibinfo{person}{Cheng Tai}, \bibinfo{person}{Tong Xiao},
  \bibinfo{person}{Yi Zhang}, \bibinfo{person}{Xiaogang Wang}, {et~al\mbox{.}}}
  \bibinfo{year}{2015}\natexlab{}.
\newblock \showarticletitle{Convolutional neural networks with low-rank
  regularization}.
\newblock \bibinfo{journal}{\emph{arXiv preprint arXiv:1511.06067}}
  (\bibinfo{year}{2015}).
\newblock


\bibitem[\protect\citeauthoryear{Tan, Tao, Ren, Tang, and Wu}{Tan
  et~al\mbox{.}}{2019}]%
        {tan2019crowd}
\bibfield{author}{\bibinfo{person}{Xin Tan}, \bibinfo{person}{Chun Tao},
  \bibinfo{person}{Tongwei Ren}, \bibinfo{person}{Jinhui Tang}, {and}
  \bibinfo{person}{Gangshan Wu}.} \bibinfo{year}{2019}\natexlab{}.
\newblock \showarticletitle{Crowd Counting via Multi-layer Regression}. In
  \bibinfo{booktitle}{\emph{Proceedings of the 27th ACM International
  Conference on Multimedia}}. \bibinfo{pages}{1907--1915}.
\newblock


\bibitem[\protect\citeauthoryear{Walach and Wolf}{Walach and Wolf}{2016}]%
        {walach2016learning}
\bibfield{author}{\bibinfo{person}{Elad Walach} {and} \bibinfo{person}{Lior
  Wolf}.} \bibinfo{year}{2016}\natexlab{}.
\newblock \showarticletitle{Learning to count with CNN boosting}. In
  \bibinfo{booktitle}{\emph{ECCV}}. Springer, \bibinfo{pages}{660--676}.
\newblock


\bibitem[\protect\citeauthoryear{Xiong, Shi, and Yeung}{Xiong
  et~al\mbox{.}}{2017}]%
        {xiong2017spatiotemporal}
\bibfield{author}{\bibinfo{person}{Feng Xiong}, \bibinfo{person}{Xingjian Shi},
  {and} \bibinfo{person}{Dit-Yan Yeung}.} \bibinfo{year}{2017}\natexlab{}.
\newblock \showarticletitle{Spatiotemporal modeling for crowd counting in
  videos}. In \bibinfo{booktitle}{\emph{ICCV}}. IEEE.
\newblock


\bibitem[\protect\citeauthoryear{Xiong, Lu, Liu, Liu, Cao, and Shen}{Xiong
  et~al\mbox{.}}{2019}]%
        {xiong2019open}
\bibfield{author}{\bibinfo{person}{Haipeng Xiong}, \bibinfo{person}{Hao Lu},
  \bibinfo{person}{Chengxin Liu}, \bibinfo{person}{Liang Liu},
  \bibinfo{person}{Zhiguo Cao}, {and} \bibinfo{person}{Chunhua Shen}.}
  \bibinfo{year}{2019}\natexlab{}.
\newblock \showarticletitle{From Open Set to Closed Set: Counting Objects by
  Spatial Divide-and-Conquer}. In \bibinfo{booktitle}{\emph{ICCV}}.
  \bibinfo{pages}{8362--8371}.
\newblock


\bibitem[\protect\citeauthoryear{Yan, Yuan, Zuo, Tan, Wang, Wen, and Ding}{Yan
  et~al\mbox{.}}{2019}]%
        {yan2019perspective}
\bibfield{author}{\bibinfo{person}{Zhaoyi Yan}, \bibinfo{person}{Yuchen Yuan},
  \bibinfo{person}{Wangmeng Zuo}, \bibinfo{person}{Xiao Tan},
  \bibinfo{person}{Yezhen Wang}, \bibinfo{person}{Shilei Wen}, {and}
  \bibinfo{person}{Errui Ding}.} \bibinfo{year}{2019}\natexlab{}.
\newblock \showarticletitle{Perspective-Guided Convolution Networks for Crowd
  Counting}. In \bibinfo{booktitle}{\emph{ICCV}}. \bibinfo{pages}{952--961}.
\newblock


\bibitem[\protect\citeauthoryear{Yim, Joo, Bae, and Kim}{Yim
  et~al\mbox{.}}{2017}]%
        {yim2017gift}
\bibfield{author}{\bibinfo{person}{Junho Yim}, \bibinfo{person}{Donggyu Joo},
  \bibinfo{person}{Jihoon Bae}, {and} \bibinfo{person}{Junmo Kim}.}
  \bibinfo{year}{2017}\natexlab{}.
\newblock \showarticletitle{A gift from knowledge distillation: Fast
  optimization, network minimization and transfer learning}. In
  \bibinfo{booktitle}{\emph{CVPR}}. \bibinfo{pages}{4133--4141}.
\newblock


\bibitem[\protect\citeauthoryear{Yuan, Qiu, Liu, Wu, Chen, Chen, and Lin}{Yuan
  et~al\mbox{.}}{2020}]%
        {yuan2020crowd}
\bibfield{author}{\bibinfo{person}{Lixian Yuan}, \bibinfo{person}{Zhilin Qiu},
  \bibinfo{person}{Lingbo Liu}, \bibinfo{person}{Hefeng Wu},
  \bibinfo{person}{Tianshui Chen}, \bibinfo{person}{Pei Chen}, {and}
  \bibinfo{person}{Liang Lin}.} \bibinfo{year}{2020}\natexlab{}.
\newblock \showarticletitle{Crowd counting via scale-communicative aggregation
  networks}.
\newblock \bibinfo{journal}{\emph{Neurocomputing}}  \bibinfo{volume}{409}
  (\bibinfo{year}{2020}), \bibinfo{pages}{420--430}.
\newblock


\bibitem[\protect\citeauthoryear{Zagoruyko and Komodakis}{Zagoruyko and
  Komodakis}{2016}]%
        {zagoruyko2016paying}
\bibfield{author}{\bibinfo{person}{Sergey Zagoruyko} {and}
  \bibinfo{person}{Nikos Komodakis}.} \bibinfo{year}{2016}\natexlab{}.
\newblock \showarticletitle{Paying more attention to attention: Improving the
  performance of convolutional neural networks via attention transfer}.
\newblock \bibinfo{journal}{\emph{arXiv preprint arXiv:1612.03928}}
  (\bibinfo{year}{2016}).
\newblock


\bibitem[\protect\citeauthoryear{Zhan, Monekosso, Remagnino, Velastin, and
  Xu}{Zhan et~al\mbox{.}}{2008}]%
        {zhan2008crowd}
\bibfield{author}{\bibinfo{person}{B Zhan}, \bibinfo{person}{Dorothy
  Monekosso}, \bibinfo{person}{Paolo Remagnino}, \bibinfo{person}{Sergio~A
  Velastin}, {and} \bibinfo{person}{Liqun Xu}.}
  \bibinfo{year}{2008}\natexlab{}.
\newblock \showarticletitle{Crowd analysis: a survey}.
\newblock \bibinfo{journal}{\emph{Machine Vision Applications}}
  \bibinfo{volume}{19}, \bibinfo{number}{5} (\bibinfo{year}{2008}),
  \bibinfo{pages}{345--357}.
\newblock


\bibitem[\protect\citeauthoryear{Zhang, Yue, Shen, Zhu, Zhen, Cao, and
  Shao}{Zhang et~al\mbox{.}}{2019a}]%
        {zhang2019attentional}
\bibfield{author}{\bibinfo{person}{Anran Zhang}, \bibinfo{person}{Lei Yue},
  \bibinfo{person}{Jiayi Shen}, \bibinfo{person}{Fan Zhu},
  \bibinfo{person}{Xiantong Zhen}, \bibinfo{person}{Xianbin Cao}, {and}
  \bibinfo{person}{Ling Shao}.} \bibinfo{year}{2019}\natexlab{a}.
\newblock \showarticletitle{Attentional Neural Fields for Crowd Counting}. In
  \bibinfo{booktitle}{\emph{ICCV}}. \bibinfo{pages}{5714--5723}.
\newblock


\bibitem[\protect\citeauthoryear{Zhang, Li, Wang, and Yang}{Zhang
  et~al\mbox{.}}{2015}]%
        {zhang2015cross}
\bibfield{author}{\bibinfo{person}{Cong Zhang}, \bibinfo{person}{Hongsheng Li},
  \bibinfo{person}{Xiaogang Wang}, {and} \bibinfo{person}{Xiaokang Yang}.}
  \bibinfo{year}{2015}\natexlab{}.
\newblock \showarticletitle{Cross-scene crowd counting via deep convolutional
  neural networks}. In \bibinfo{booktitle}{\emph{CVPR}}.
  \bibinfo{pages}{833--841}.
\newblock


\bibitem[\protect\citeauthoryear{Zhang, Zhu, and Ye}{Zhang
  et~al\mbox{.}}{2019b}]%
        {zhang2019fast}
\bibfield{author}{\bibinfo{person}{Feng Zhang}, \bibinfo{person}{Xiatian Zhu},
  {and} \bibinfo{person}{Mao Ye}.} \bibinfo{year}{2019}\natexlab{b}.
\newblock \showarticletitle{Fast Human Pose Estimation}. In
  \bibinfo{booktitle}{\emph{CVPR}}.
\newblock


\bibitem[\protect\citeauthoryear{Zhang, Wu, Costeira, and Moura}{Zhang
  et~al\mbox{.}}{2017}]%
        {zhang2017fcn}
\bibfield{author}{\bibinfo{person}{Shanghang Zhang}, \bibinfo{person}{Guanhang
  Wu}, \bibinfo{person}{Joao~P Costeira}, {and} \bibinfo{person}{Jos{\'e}~MF
  Moura}.} \bibinfo{year}{2017}\natexlab{}.
\newblock \showarticletitle{Fcn-rlstm: Deep spatio-temporal neural networks for
  vehicle counting in city cameras}. In \bibinfo{booktitle}{\emph{ICCV}}. IEEE,
  \bibinfo{pages}{3687--3696}.
\newblock


\bibitem[\protect\citeauthoryear{Zhang, Xiang, Hospedales, and Lu}{Zhang
  et~al\mbox{.}}{2018}]%
        {zhang2018deep}
\bibfield{author}{\bibinfo{person}{Ying Zhang}, \bibinfo{person}{Tao Xiang},
  \bibinfo{person}{Timothy~M Hospedales}, {and} \bibinfo{person}{Huchuan Lu}.}
  \bibinfo{year}{2018}\natexlab{}.
\newblock \showarticletitle{Deep mutual learning}. In
  \bibinfo{booktitle}{\emph{CVPR}}. \bibinfo{pages}{4320--4328}.
\newblock


\bibitem[\protect\citeauthoryear{Zhang, Zhou, Chen, Gao, and Ma}{Zhang
  et~al\mbox{.}}{2016}]%
        {zhang2016single}
\bibfield{author}{\bibinfo{person}{Yingying Zhang}, \bibinfo{person}{Desen
  Zhou}, \bibinfo{person}{Siqin Chen}, \bibinfo{person}{Shenghua Gao}, {and}
  \bibinfo{person}{Yi Ma}.} \bibinfo{year}{2016}\natexlab{}.
\newblock \showarticletitle{Single-image crowd counting via multi-column
  convolutional neural network}. In \bibinfo{booktitle}{\emph{CVPR}}.
  \bibinfo{pages}{589--597}.
\newblock


\bibitem[\protect\citeauthoryear{Zhou, Wu, Ni, Zhou, Wen, and Zou}{Zhou
  et~al\mbox{.}}{2016}]%
        {zhou2016dorefa}
\bibfield{author}{\bibinfo{person}{Shuchang Zhou}, \bibinfo{person}{Yuxin Wu},
  \bibinfo{person}{Zekun Ni}, \bibinfo{person}{Xinyu Zhou}, \bibinfo{person}{He
  Wen}, {and} \bibinfo{person}{Yuheng Zou}.} \bibinfo{year}{2016}\natexlab{}.
\newblock \showarticletitle{Dorefa-net: Training low bitwidth convolutional
  neural networks with low bitwidth gradients}.
\newblock \bibinfo{journal}{\emph{arXiv preprint arXiv:1606.06160}}
  (\bibinfo{year}{2016}).
\newblock


\bibitem[\protect\citeauthoryear{Zhu and Gupta}{Zhu and Gupta}{2017}]%
        {zhu2017prune}
\bibfield{author}{\bibinfo{person}{Michael Zhu} {and} \bibinfo{person}{Suyog
  Gupta}.} \bibinfo{year}{2017}\natexlab{}.
\newblock \showarticletitle{To prune, or not to prune: exploring the efficacy
  of pruning for model compression}.
\newblock \bibinfo{journal}{\emph{arXiv preprint arXiv:1710.01878}}
  (\bibinfo{year}{2017}).
\newblock


\end{thebibliography}

%
%
%
%

\end{document}